\documentclass[]{article}

\usepackage{chngpage}
\usepackage{mkolar_definitions}
\usepackage{rotating}
\usepackage{multirow}

\RequirePackage{natbib}

\usepackage{algorithm}
\usepackage{algorithmic}
\usepackage{graphics}
\usepackage{subfigure}

\bibpunct{(}{)}{;}{a}{,}{,}

\title{Ultra-high Dimensional
  Multiple Output Learning With Simultaneous Orthogonal Matching
  Pursuit: A Sure Screening Approach}

\author{Mladen Kolar and Eric P. Xing\\
School of Computer Science\\
Carnegie Mellon University}

\begin{document}

\maketitle

\begin{abstract}
  We propose a novel application of the Simultaneous Orthogonal
  Matching Pursuit (S-OMP) procedure for sparsistant variable
  selection in ultra-high dimensional multi-task regression problems.
  Screening of variables, as introduced in \cite{fan08sis}, is an
  efficient and highly scalable way to remove many irrelevant
  variables from the set of all variables, while retaining all the
  relevant variables. S-OMP can be applied to problems with hundreds
  of thousands of variables and once the number of variables is
  reduced to a manageable size, a more computationally demanding
  procedure can be used to identify the relevant variables for each of
  the regression outputs.  To our knowledge, this is the first attempt
  to utilize relatedness of multiple outputs to perform fast screening
  of relevant variables. As our main theoretical contribution, we
  prove that, asymptotically, S-OMP is guaranteed to reduce an
  ultra-high number of variables to below the sample size without
  losing true relevant variables.  We also provide formal evidence
  that a modified Bayesian information criterion (BIC) can be used to
  efficiently determine the number of iterations in S-OMP.  We further
  provide empirical evidence on the benefit of variable selection
  using multiple regression outputs jointly, as opposed to performing
  variable selection for each output separately.  The finite sample
  performance of S-OMP is demonstrated on extensive simulation
  studies, and on a genetic association mapping problem. 

\end{abstract}

\section{Introduction}

Multiple output regression, also known as multi-task regression, with
{\it ultra-high dimensional} inputs commonly arise in problems such as
genome-wide association (GWA) mapping in genetics, or stock portfolio
prediction in finance. For example, in a GWA mapping problem, the goal
is to find a small set of relevant single-nucleotide polymorphisms
(SNP) ({\it covariates, or inputs}) that account for variations of a
large number of gene expression or clinical traits ({\it responses, or
  outputs}), through a response function that is often modeled via a
regression. However, this is a very challenging problem for current
statistical methods since the number of input variables is likely to
reach millions, prohibiting even usage of scalable implementation of
Lasso-like procedures for model selection, which are a convex
relaxation of a combinatorial subset selection search. Furthermore,
the outputs in a typical multi-task regression problem are not
independent of each other, therefore the discovery of truly relevant
inputs has to take into consideration of potential joint effects
induced by coupled responses. To appreciate this better, consider
again the GWA example. Typically, genes in a biological pathway are
co-expressed as a module and it is often assumed that a causal SNP
affects multiple genes in one pathway, but not all of the genes in the
pathway. In order to effectively reduce the dimensionality of the
problem and to detect the causal SNPs, it is very important to look at
how SNPs affect all genes in a biological pathway.  Since the
experimentally collected data is usually very noisy, regressing genes
individually onto SNPs may not be sufficient to identify the relevant
SNPs that are only weakly marginally correlated with each individual
gene in a module. However, once the whole biological pathway is
examined, it is much easier to find such causal SNPs. In this paper,
we demonstrate that the Simultaneous Orthogonal Matching Pursuit
(S-OMP) \citep{tropp06algorithms_part_1} can be used to quickly reduce
the dimensionality of such problems, without losing any of the
relevant variables.

From a computational point of view, as the dimensionality of the
problem and the number of outputs increase, it can become intractable
to solve the underlying convex programs commonly used to identify
relevant variables in multi-task regression problems. Previous work by
\cite{han09blockwise}, \cite{lounici09taking} and
\cite{kim09multivariate}, for example, do not scale well to settings
when the number of variables exceeds $\gtrsim 10000$ and the number of
outputs exceeds $\gtrsim 1000$, as in typical genome-wide association
studies. Furthermore, since the estimation error of the regression
coefficients depends on the number of variables in the problem,
variable selection can improve convergence rates of estimation
procedures. These concerns motivate us to propose and study the S-OMP
as a fast way to remove irrelevant variables from an ultra-high
dimensional space.

Formally, the GWA mapping problem, which we will use as an
illustrative example both in here for model formulation and later for
empirical experimental validation, can be cast as a variable selection
problem in a multiple output regression model:
\begin{equation} \label{eq:model_linear2}
\Yb = \Xb\Bb + \Wb
\end{equation}
where $\Yb = [\yb_1, \ldots, \yb_T] \in \RR^{n \times T}$ is a matrix
of outputs, whose column $\yb_t$ is an $n$-vector for the $t$-th
output (i.e., gene), $\Xb \in \RR^{n \times p}$ is a random design
matrix, of which each row $\xb_i$ denotes a $p$-dimensional input,
$\Bb = [\betab_1, \ldots, \betab_T] \in \RR^{p \times T}$ is the
matrix of regression coefficients and $\Wb = [\epsilon_1, \ldots,
\epsilon_T] \in \RR^{n \times T}$ is a matrix of IID random noise,
independent of $\Xb$. Throughout the paper we are going to assume that
the columns of $\Bb$ are jointly sparse, as we precisely specify
below. Note that if different columns of $\Bb$ do not share any
underlying structure, the model in \eqref{eq:model_linear2} can be
estimated by fitting each of the tasks separately.

We are interested in estimating the regression coefficients, under the
assumption that they share a common structure, e.g., there exist a
subset of variables with non-zero coefficients for more than one
regression output. We informally refer to such outputs as related.
Such a variable selection problem can be formalized in two ways: (1)
the {\it union support} recovery of $\Bb$, as defined in
\cite{obozinski10support}, where a subset of variables is selected
that affect at least one output; (2) the {\it exact support} recovery
of $\Bb$, where the exact positions of non-zero elements in $\Bb$ are
estimated. In this paper, we concern ourselves with exact support
recovery, which is of particular importance in problems like GWA
mapping \citep{kim09statistical} or biological network estimation
\citep{peng08regularized}. Under such a multi-task setting, two
interesting questions naturally follow: i) how can information be
shared between related outputs in order to improve the predictive
accuracy and the rate of convergence of the estimated regression
coefficients over the independent estimation on each output
separately; ii) how to select relevant variables more accurately based
on information from related outputs.  To address these two questions,
one line of research \citep[e.g.,][]{zhang06probabilistic,
  han09blockwise, lounici09taking} has looked into the following
estimation procedure leveraging a {\it multi-task regularization}:
\begin{equation} \label{eq:group_optimization}
\hat \Bb = \argmin_{\betab_t \in \RR^{p}, t \in [T]} \sum_{t = 1}^{T} 
\norm{\yb_t - \Xb\betab_t}_2^2 + \lambda \sum_{j = 1}^p {\rm
  pen}(\beta_{1,j}, \ldots, \beta_{T,j}),
\end{equation}
with ${\rm pen}(a_1, \ldots, a_T) = \max_{t \in [T]} |a_t|$ or ${\rm
  pen}(a_1, \ldots, a_T) = \sqrt{\sum_{t \in [T]} a_t^2}$ for a vector
$\ab \in \RR^{T}$. Under an appropriate choice of the penalty
parameter $\lambda$, the estimator $\hat \Bb$ has many rows equal to
zero, which correspond to irrelevant variables. However, solving
\eqref{eq:group_optimization} can be computationally prohibitive.

In this paper, we consider an ultra-high dimensional setting for the
aforementioned multi-task regression problem, where the number of
variables $p$ is much higher than the sample size $n$, e.g. $p =
\Ocal(\exp(n^{\delta_p}))$ for a positive constant $\delta_p$, but the
regression coefficients $\betab_t$ are sparse, i.e., for each task
$t$, there exist a very small number of variables that are relevant to
the output. Under the sparsity assumption, it is highly important to
efficiently select the relevant variables in order to improve the
accuracy of the estimation and prediction, and to facilitate the
understanding of the underlying phenomenon for domain experts. In the
seminal paper of \cite{fan08sis}, the concept of {\it sure screening}
was introduced, which leads to a sequential variable selection
procedure that keeps all the relevant variables with high probability
in ultra-high dimensional {\it uni-output regression}. In this paper,
we propose the S-OMP procedure, which enjoys {\it sure screening}
property in ultra-high dimensional {\it multiple output regression} as
defined in \eqref{eq:model_linear2}. To perform {\it exact support}
recovery, we further propose a two-step procedure that first use S-OMP
to screen the variables, i.e., select a subset of variables that
contain all the true variables; and then use the adaptive Lasso
(ALasso) \citep{zou06adaptive} to further select a subset of screened
variables for each task. We show, both theoretically and empirically,
that our procedure ensure sparsistant recovery of relevant
variables. To the best of our knowledge, this is the first attempt to
analyze the sure screening property in the ultra-high dimensional
space using the shared information from the multiple regression
outputs.

\subsection{Related Work}
The model given in \eqref{eq:model_linear2} has been used in many
different domains ranging from multivariate regression
\citep{obozinski09high, negahban09Phase} and sparse approximation
\citep{tropp06algorithms_part_1} to neural science
\citep{han09blockwise}, multi-task learning \citep{lounici09taking,
  Argyriou08convex} and biological network estimation
\citep{peng08regularized}. A number of authors has provided
theoretical understanding of the estimation in the model using the
convex program \eqref{eq:group_optimization} to estimate $\hat \Bb$.
\cite{lounici09taking} showed the benefits of the joint estimation,
when there is a small set of variables common to all outputs and the
number of outputs is large. \cite{obozinski09high} and
\cite{negahban09Phase} analyzed the consistent recovery of the union
support. \cite{negahban09Phase} provided the analysis of the exact
support recovery for a special case with two outputs.

The Orthogonal Matching Pursuit (OMP) has been analyzed before in the
literature \citep[see, e.g.,][]{zhang09consistency, lozano09grouped,
  wang09forward, barron08approximation}. In particular, our work
should be contrasted to \cite{wang09forward}, which showed that the
OMP has the sure screening property in a linear regression with a
single output, and to the exact variable selection property of the OMP
analyzed in \cite{zhang09consistency} and \cite{lozano09grouped}. The
exact variable selection requires much stronger assumptions on the
design, such as the irrepresentable condition, that are hard to
satisfy in the ultra-high dimensional setting. On the other hand, the
sure screening property can be shown to hold under much weaker
assumptions. 

In this paper, we make the following novel contributions: i) we prove
that the S-OMP can be used for the ultra-high dimensional variable
screening in multiple output regression problems and demonstrate its
performance on extensive numerical studies; ii) we show that a two
step procedure can be used to select exactly the relevant variables
for each task; and iii) we prove that a modification of the BIC score
\citep{Chen08extended} can be used to select the number of steps in the
S-OMP.

The rest of the article is organized as follows. In Section 2, we
introduce the simultaneous greedy forward regression and propose our
approach to the exact support estimation. Theoretical guarantees of
the methods are given in Section 3. Section 4 is devoted to
extensive numerical simulations. An application to the real world
problem in association mapping is demonstrated in Section 5. We
conclude with discussion in Section 6. Proofs are deferred to
Appendix. 

\section{Methodology}

\subsection{The model and notation}

We will consider a slightly more general model 
\begin{equation} \label{eq:model_linear}
  \begin{aligned}
    \yb_1 &= \Xb_1 \betab_1 + \epsilonb_1 \\
    \yb_2 &= \Xb_2 \betab_2 + \epsilonb_2 \\
    & \ldots\\
    \yb_T &= \Xb_T \betab_T + \epsilonb_T,
  \end{aligned}
\end{equation}
than the one given in \eqref{eq:model_linear2}.  The model in
\eqref{eq:model_linear2} is a special case of the model in
\eqref{eq:model_linear}, with all the design matrices $\{\Xb_t\}_{t
  \in [T]}$ being equal and $[T]$ denoting the set $\{1, \ldots, T\}$.
Assume that for all $t \in [T]$, $\Xb_t \in \mathbb{R}^{n \times
  p}$. For the design $\Xb_t$, we denote $\Xb_{t,j}$ the $j$-th column
(i.e., dimension), $\xb_{t,i}$ the $i$-th row (i.e., instance) and
$x_{t,ij}$ the element at $(i,j)$. Denote $\Sigmab_t =
\Cov(\xb_{t,i})$. Without loss of generality, we assume that
$\Var(y_{t,i}) = 1$, $\EE(x_{t,ij}) = 0$ and $\Var(x_{t, ij}) = 1$.
The noise $\epsilonb_{t}$ is zero mean and $\Cov(\epsilonb_t) =
\sigma^2\Ib_{n \times n}$.  We assume that the number of variables $p
\gg n$ and that the vector of regression coefficients $\betab_t$ are
jointly sparse, that is, there exist a small number of variables that
are relevant for the most of the $T$ regression problems. Put another
way, the matrix $B = [\betab_1, \ldots, \betab_T]$ has only a small
number of non-zero rows.  Let $\Mcal_{*,t}$ denote the set of non-zero
coefficients of $\betab_t$ and $\Mcal_* = \cup_{t = 1}^T \Mcal_{*,t}$
denote the set of all relevant variables, i.e., variables with
non-zero coefficient in at least one of the tasks. For an arbitrary
set $\Mcal \subseteq \{1, \ldots, p\}$, $\Xb_{t, \Mcal}$ denotes the
design with columns indexed by $\Mcal$, $\Bb_{\Mcal}$ denotes the rows
of $\Bb$ indexed by $\Mcal$ and $\Bb_j = (\betab_{1,j}, \ldots,
\betab_{T, j})'$. The cardinality of the set $\Mcal$ is denoted as
$|\Mcal|$. Let $s := |\Mcal_*|$ denote the total number of relevant
variables, so under the sparsity assumption we have $s < n$. For a
square matrix $\Ab$, $\Lambda_{\min}(\Ab)$ and $\Lambda_{\max}(\Ab)$
are used to denote the minimum and the maximum eigenvalue,
respectively. For a different matrix $\Ab = [a_{ij}] \in \RR^{p \times
  T}$, we define $\norm{\Ab}_{2,1} := \sum_{i \in [p]} \sqrt{\sum_{j
    \in [T]} a_{ij}^2}$. Lastly, we use $[p]$ to denote the set $\{1,
\ldots, p\}$.

\subsection{Simultaneous Orthogonal Matching Pursuit}

\begin{algorithm}[t]
\caption{Group Forward Regression}
{\small
\textbf{Input}: Dataset $\{\Xb_t, \yb_t\}_{t=1}^T$\\
\textbf{Output}: Sequence of selected models $\{\Mcal^{(k)}\}_{k=0}^{n-1}$\\[-0.3cm]
\begin{algorithmic}[1]
\STATE Set $\Mcal^{(0)} = \emptyset$ \\ 
\FOR{$k=1$ to $n-1$}
  \FOR{$j=1$ to $p$}
  \STATE $\tilde{\Mcal}_j^{(k)} = \Mcal^{(k-1)} \cup \{j\}$ \\
  \STATE $\Hb_{t,j} = \Xb_{t, \tilde{\Mcal}_j^{(k)}} (\Xb_{t, \tilde{\Mcal}_j^{(k)}}'
  \Xb_{t, \tilde{\Mcal}_j^{(k)}})^{-1} \Xb_{t, \tilde{\Mcal}_j^{(k)}}'$ \\
  \STATE ${\rm RSS}(\tilde{\Mcal}_j^{(k)}) = \sum_{t=1}^T \yb_t'(\Ib_{n \times n} -
  \Hb_{t,j})\yb_t$ \\
  \ENDFOR
  \STATE $\hat f_k = \argmin_{j \in \{1, \ldots, p\} \backslash
    \Mcal^{(k-1)}} {\rm RSS}(\tilde{\Mcal}_j^{(k)})$ \\
  \STATE $\Mcal^{(k)} = \Mcal^{(k-1)} \cup \{\hat f_k\}$ \\
\ENDFOR
\end{algorithmic}
\label{alg:group_forward}
}
\end{algorithm}

We propose a Simultaneous Orthogonal Matching Pursuit procedure for
ultra high-dimensional variable selection in the multi-task regression
problem, which is outlined in Algorithm~\ref{alg:group_forward}.
Before describing the algorithm, we introduce some additional
notation. For an arbitrary subset $\Mcal \subseteq [p]$ of variables,
let $\Hb_{t, \Mcal}$ be the orthogonal projection matrix onto ${\rm
  Span}( \Xb_{t, \Mcal} )$, i.e.,
\begin{equation}
  \Hb_{t, \Mcal} = \Xb_{t, \Mcal} (\Xb_{t, \Mcal}'
  \Xb_{t, \Mcal})^{-1} \Xb_{t, \Mcal}',
\end{equation}
and define the residual sum of squares (RSS) as
\begin{equation}
{\rm RSS}(\Mcal) = \sum_{t=1}^T \yb_t'(\Ib_{n \times n} -
\Hb_{\Mcal})\yb_t.
\end{equation}

The algorithm starts with an empty set $\Mcal^{(0)} = \emptyset$. We
recursively define the set $\Mcal^{(k)}$ based on the set
$\Mcal^{(k-1)}$. The set $\Mcal^{(k)}$ is obtained by adding a
variable indexed by $\hat f_k \in [p]$, which minimizes ${\rm
  RSS}(\Mcal^{(k-1)} \cup j)$ over the set $[p] \backslash
\Mcal^{(k-1)}$, to the set $\Mcal^{(k-1)}$. Repeating the algorithm
for $n-1$ steps, a sequence of nested sets $\{ \Mcal^{(k)}\}_{k =
  0}^{n-1}$ is obtained, with $\Mcal^{(k)} = \{\hat f_1, \ldots, \hat
f_k\}$.

To practically select one of the sets of variables from $\{
\Mcal^{(k)}\}_{k = 0}^{n-1}$, we minimize the modified BIC criterion
\citep{Chen08extended}, which is defined as
\begin{equation}
  \label{eq:bic_criterion}
  {\rm BIC}(\Mcal) = \log \left( \frac{{\rm RSS}(\Mcal)}{nT} \right) +
  \frac{|\Mcal|(\log(n) + 2\log(p))}{n}
\end{equation}
with $|\Mcal|$ denoting the number of elements of the set~$\Mcal$.
Let 
$$\hat s = \argmin_{k \in \{0, \ldots, n-1\}} {\rm BIC}
(\Mcal^{(k)}),$$ so that the selected model is $\Mcal^{(\hat s)}$.

{\bf Remark:} The S-OMP algorithm is outlined only conceptually in
this section. The steps 5 and 6 of the algorithm can be implemented
efficiently using the progressive Cholesky decomposition see, e.g.,
\cite{cotter99forward}. A computationally costly step 5 involves
inversion of the matrix $\Xb_{t, \Mcal}' \Xb_{t, \Mcal}$, however, it
can be seen from the algorithm that the matrix $\Xb_{t, \Mcal}'
\Xb_{t, \Mcal}$ is updated in each iteration by simply appending a row
and a column to it. Therefore, its Cholesky factorization can be
efficiently computed based on calculation involving only the last
row. A detailed implementation of the orthogonal matching pursuit
algorithm based on the progressive Cholesky decomposition can be found
in \cite{rubinstein08efficient}.

\subsection{Exact variable selection}

\begin{figure}[!hb]
  \centering
  \includegraphics[width=0.9\textwidth]{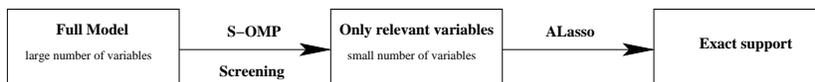}
  \caption{Framework for exact suppor recovery}
  \label{fig:framework}
\end{figure}

After removing many of the irrelevant variables have been removed
using Algorithm~\ref{alg:group_forward}, we are left with the
variables in the set $\Mcal^{(\hat s)}$, whose size is smaller than
the sample size $n$. These variables are candidates for the relevant
variables for each of the regressions. Now, we can address the problem
of estimating the regression coefficients and recovering the exact
support of $\Bb$ using a lower dimensional selection procedure. In
this paper, we use the adaptive Lasso as a lower dimensional selection
procedure, which was shown to have oracle properties
\citep{zou06adaptive}. The ALasso solves the penalized least square
problem
\begin{equation} \label{eq:adaptive_lasso}
\hat \betab_t = \argmin_{\betab_t \in \RR^{\hat s}}
\norm{\yb_t - \Xb_{t, \Mcal^{(\hat s)}}\betab_t}_2^2 + 
\lambda \sum_{j \in \Mcal^{(\hat s)}} w_j |\beta_{t,j}|, 
\end{equation}
where $(w_j)_{j \in \Mcal^{(\hat s)}}$ is a vector of known weight and
$\lambda$ is a tuning parameter. Usually, the weights are defined as
$w_j = 1/|\hat \beta_{t,j}|$ where $\hat \betab_t$ is a
$\sqrt{n}$-consistent estimator of $\betab_t$. In a low dimensional
setting, we know from \cite{huang08adaptive} that the adaptive Lasso
can recover the exactly the relevant variables. Therefore, we can use
the ALasso on each output separately to recover the exact support of
$\Bb$. However, in order to ensure that the exact support of $\Bb$ is
recovered with high probability, we need to have that the total number
of tasks is $o(n)$. The exact support recovery of $\Bb$ is established
using the union bound over different tasks, therefore we need the
number of tasks to remain relatively small in comparison to the sample
size $n$. However, simulation results presented in
$\S$~ref{sec:simulation} show that the ALasso procedure succeeds in
the exact support recovery even when the number of tasks are much
larger than the sample size, which indicates that our theoretical
considerations could be improved. Figure~\ref{fig:framework}
illustrates the two step procedure.

 {\bf Remark:} We point out that solving the multi-task
problem defined in \eqref{eq:group_optimization} can be efficiently
done on the reduced set of variables, but it is not obvious how to
obtain the estimate of the exact support using
\eqref{eq:group_optimization}. In Section~\ref{sec:simulation}, our
numerical studies show that the ALasso applied to the reduced set of
variables can be used to estimate the exact support of $\Bb$.

\section{Theory}

In this section we state conditions under which
Algorithm~\ref{alg:group_forward} is screening consistent, i.e.,
\begin{equation}
  \PP[\exists k \in \{0, 1, \ldots, n-1\} : \Mcal_* \subseteq
  \Mcal^{(k)}] \rightarrow 1,\ \text{as } n \rightarrow \infty.
\end{equation}
Furthermore, we also show that the model selected using the modified
BIC criterion contains all the relevant variables, i.e., 
\begin{equation}
  \PP[\Mcal_* \subseteq \Mcal^{(\hat s)}] \rightarrow 1,\ \text{as } n \rightarrow \infty.
\end{equation}
Note that we can choose trivially $\Mcal^{(n)}$ since it holds that
$\Mcal_* \subseteq \Mcal^{(n)}$. However, we will be able to prove
that $\hat s$ chosen by the modified BIC criterion is much smaller
than the sample size $n$.

\subsection{Assumptions}

Before we state the theorem characterizing the performance of the
S-OMP, we give some technical conditions that are needed for our
analysis.

\begin{description}
\item{\bf A1:} The random noise vectors $\epsilonb_1, \ldots, \epsilonb_T$
  are independent Gaussian with zero mean and covariance matrix
  $\sigma^2\Ib_{n \times n}$.
\item{\bf A2:} Each row of the design matrix $\Xb_t$ is IID Gaussian
  with zero mean and covariance matrix $\Sigmab_t$. Furthermore, there
  exist two positive constants $0 < \phi_{\min} < \phi_{\max} <
  \infty$ such that
  \begin{equation}
    \phi_{\min} \leq \min_{t \in [T]} \Lambda_{\min}(\Sigmab_t) 
               \leq \max_{t \in [T]} \Lambda_{\max}(\Sigmab_t)
               \leq \phi_{\max}.
  \end{equation}
\item{\bf A3:} The true regression coefficients are bounded, i.e.,
  there exists a positive constant $C_{\betab}$ such that
  $\norm{\Bb}_{2,1} \leq C_{\betab}$. Furthermore, the norm of any
  non-zero row of the matrix $\Bb$ is bounded away from zero, that is,
  there exist positive constants $c_{\betab}$ and $\delta_{\min}$ such
  that 
  \begin{equation*}
    T^{-1} \min_{j \in \Mcal_*} \sum_{t \in [T]} \beta_{t, j}^2
    \geq c_{\betab} n^{-\delta_{\min}}.    
  \end{equation*}
\item{\bf A4:} There exist positive constants $C_{s}, C_{p}, \delta_s$
  and $\delta_p$ such that $|\Mcal_*| \leq C_sn^{\delta_s}$ and
  $\log(p) \leq C_p n^{\delta_p}$.
\end{description}
The normality condition {\bf A1} is assumed here only to facilitate
presentation of theoretical results, as is commonly assumed in
literature, \cite[e.g.,][]{zhang08sparsity, fan08sis}. The normality
assumption can be avoided at the cost of more technical proofs, e.g.,
\cite{lounici09taking}, where the main technical difficulty is showing
that the concentration properties still hold. Under the condition {\bf
  A2} we will be able to show that the empirical covariance matrix
satisfies the sparse eigenvalue condition (see
Lemma~\ref{lem:sparse_eigenvalue}) with probability tending to one.
The assumption that the rows of the design are Gaussian can be easily
relaxed to the case when the rows are sub-Gaussian, without any
technical difficulties in proofs, since we would still obtain
exponential bounds on the tail probabilities. The condition {\bf A3}
states that the regression coefficients are bounded, which is a
technical condition likely to be satisfied in practice. Furthermore,
it is assumed that the row norms of $\Bb_{\Mcal_*}$ do not decay to
zero too fast or, otherwise, they would not be distinguishable from
noise. The condition is not too restrictive, e.g., if every non-zero
coefficient is bounded away from zero by a constant, the condition
{\bf A3} is trivially satisfied with $\delta_{\min} = 0$.  However, we
allow the coefficients of the relevant variables to get smaller as the
sample size increases and still guarantee that the relevant variable
will be identified. The condition {\bf A4} sets the upper bound on the
number of relevant variables and the total number of variables. While
the total number of variables can diverge to infinity much faster than
the sample size, the number of relevant variables needs to be smaller
than the sample size. Conditions {\bf A3} and {\bf A4} implicitly
relate different outputs and control the number of non-zero
coefficients shared between different outputs. Intuitively, if the
upper bound in {\bf A4} on the size of $\Mcal_*$ is large, this
immediately implies that the constant $C_{\beta}$ in {\bf A3} should
be large as well, since otherwise there would exist a row of $\Bb$
whose $\ell_2$ norm would be too small to be detected by
Algorithm~\ref{alg:group_forward}.

\subsection{Screening consistency}

Our first results states that after a small number of iterations,
compared to the dimensionality $p$, the S-OMP procedure will include
all the relevant variables. 

\begin{theorem} \label{thm:path_screening_consistent}
Assume the model in \eqref{eq:model_linear} and that
the conditions {\bf A1}-{\bf A4} are satisfied. Furthermore, assume
that 
\begin{equation}
\frac{n^{1-6\delta_s-6\delta_{\min}}}{\max\{\log(p),\log(T)\}}
\rightarrow \infty,\text{ as } n \rightarrow \infty.
\end{equation}
Then there exists a number $m_{\max}^* = m_{\max}^*(n)$, so that in
$m_{\max}^*$ steps of S-OMP iteration, all the relevant variables are
included in the model, i.e., as $n \rightarrow \infty$
\begin{equation}
  \PP[\Mcal_* \subseteq \Mcal^{(m_{\max}^*)}] \geq 1 - C_1\exp\left(-C_2
  \frac{n^{1-6\delta_s-6\delta_{\min}}} {\max\{\log(p), \log(T)\}}\right),
\end{equation}
for some positive constants $C_1$ and $C_2$.
The exact value of $m_{\max}^*$ is given as 
\begin{equation} \label{eq:max_num_steps}
m_{\max}^* = \lfloor 2^{4}\phi_{\min}^{-2}\phi_{\max}C_{\betab}^{2}
  C_s^2c_{\betab}^{-2} n^{2\delta_s + 2\delta_{\min}} \rfloor.
\end{equation}
\end{theorem} 
\noindent{\bf Remarks:} Under the assumptions of
Theorem~\ref{thm:path_screening_consistent}, $m_{\max}^* \leq n-1$, so
that the procedure effectively reduces the dimensionality below the
sample size. From the proof of the theorem, it is clear how multiple
outputs help to identify the relevant variables. The crucial quantity
in identifying all relevant variables is the minimum non-zero row norm
of $\Bb$, which allows us to identify weak variables if they are
relevant for a large number of outputs even though individual
coefficients may be small. It should be noted that the main
improvement over the ordinary forward regression is in the seize of
the signal that can be detected, as defined in {\bf A3} and {\bf A4}.

Theorem~\ref{thm:path_screening_consistent} guarantees that one of the
sets $\{\Mcal^{(k)}\}$ will contain all relevant variables, with high
probability. However, it is of practical importance to select of one
set in the collection that contains all relevant variables and does
not have too many irrelevant ones. Our following theorem shows that
the modified BIC criterion can be used for this purpose, that is, 
the set $\Mcal^{(\hat s)}$  is screening consistent. 

\begin{theorem} \label{thm:bic_screening_consistent}
Assume that the conditions of
Theorem~\ref{thm:path_screening_consistent} are satisfied. Let
\begin{equation} 
  \hat s = \argmin_{k \in \{0, \ldots, n-1\}} {\rm BIC}(\Mcal^{(k)})
\end{equation}
be the index of the model selected by optimizing the modified BIC
criterion.  Then, as $n \rightarrow \infty$
\begin{equation}
  \PP[\Mcal_* \subseteq \Mcal^{(\hat s)}] \rightarrow 1.
\end{equation}
\end{theorem}

Combining results from Theorem~\ref{thm:path_screening_consistent} and
Theorem~\ref{thm:bic_screening_consistent}, we have shown that the
S-OMP procedure is screening consistent and can be applied to problems
where the dimensionality of the problem $p$ is exponential in the
number of observed samples. In the next section, we also show that the
S-OMP has great empirical performance.

\section{Numerical studies}

In this section we perform simulation studies on an extensive number
synthetic data sets. Furthermore, we demonstrate the application of
the procedure on the genome-wide association mapping problem.

\subsection{Simulation studies} \label{sec:simulation}

We conduct an extensive number of numerical studies to evaluate the
finite sample performance of the S-OMP. We consider three procedures
that perform estimation on individuals outputs: Sure Independence
Screening (SIS), Iterative SIS (ISIS) \citep{fan08sis}, and the OMP,
for comparison purposes. The evaluation is done on the model in
\eqref{eq:model_linear2}. SIS and ISIS are used to select a subset of
variables and then the ALasso is used to further refine the
selection. We denote this combination as SIS-ALasso and
ISIS-ALasso. The size of the model selected by SIS is fixed as $n-1$,
while the ISIS selects $\lfloor n/\log(n)\rfloor$ variables in each of
the $\lfloor \log(n) - 1 \rfloor$ iterations. From the screened
variables, the final model is selected using the ALasso, together with
the BIC criterion \eqref{eq:bic_criterion} to determine the penalty
parameter $\lambda$. The number of variables selected by the OMP is
determined using the BIC criterion, however, we do not further refine
the selected variables using the ALasso, since from the numerical
studies in \cite{wang09forward} it was observed that the further
refinement does not result in improvement. The S-OMP is used to reduce
the dimensionality below the sample size jointly using the regression
outputs. Next, the ALasso is used on each of the outputs to further
perform the estimation. This combination is denoted SOMP-ALasso.

Let $\hat \Bb = [\hat \betab_1, \ldots, \hat \betab_T] \in \RR^{p
  \times T}$ be an estimate obtained by one of the estimation
procedures. We evaluate the performance averaged over 200 simulation
runs. Let $\hat \EE_n$ denote the empirical average over the
simulation runs. We measure the size of the union support $\hat S =
S(\hat \Bb) := \{j \in [p] : \norm{ \hat \Bb_j}_2^2 > 0 \}$. Next, we
estimate the probability that the screening property is satisfied
$\hat \EE_n[ \ind \{\Mcal_* \subseteq S(\hat \Bb)\} ]$, which we call
coverage probability. For the union support, we define fraction of
correct zeros $(p-s)^{-1} \hat\EE_n[|S(\hat \Bb)^C \cap \Mcal_*^C|]$,
fraction of incorrect zeros $s^{-1} \hat\EE_n[|S(\hat \Bb)^C \cap
\Mcal_*|]$ and fraction of correctly fitted $\hat\EE_n[\ind \{\Mcal_*
= S(\hat \Bb)\}]$ to measure the performance of different procedures.
Similar quantities are defined for the exact support recovery. In
addition, we measure the estimation error $\hat \EE_n[\norm{\Bb - \hat
  \Bb}_2^2]$ and the prediction performance on the test set. On the
test data $\{\xb_i^*, \yb_i^*\}_{i \in [n]}$, we compute
\begin{equation}
R^2 = 1 - \frac{\sum_{i \in [n]} \sum_{t \in [T]} (y_{t,i}^* -
    (\xb_{t,i}^*)'\hat \betab_{t})^2}{\sum_{i \in [n]} \sum_{t \in
      [T]} (y_{t,i}^* - \bar{y_t^*})^2},
\end{equation}
where $\bar{y_t^*} = n^{-1}\sum_{i \in [n]}y_{t,i}$.

The following simulation studies are used to comparatively assess the
numerical performance of the procedures. Due to space constraints,
tables with detailed numerical results are given in the appendix. In
this section, we outline main findings.

{\it Simulation 1:} [Model with uncorrelated variables] The following
toy model is based on the simulation I in \cite{fan08sis} with $(n, p,
s, T) = (400, 20000, 18, 500)$. Each $\xb_{i}$ is drawn independently
from a standard multivariate normal distribution, so that the
variables are mutually independent. For $j \in [s]$ and $t \in [T]$,
the non-zero coefficients of $\Bb$ are given as $\beta_{t, j} =
(-1)^{u}(4n^{-1/2}\log n + |z|)$, where $u \sim {\rm Bernoulli}(0.4)$
and $z \sim \Ncal(0, 1)$. The number of non-zero elements in $\Bb_j$
is given as a parameter $T_{\rm non-zero} \in \{500, 300, 100\}$. The
positions of non-zero elements are chosen uniformly at random from
$[T]$. The noise is Gaussian with the standard deviation $\sigma$ set
to control the signal-to-noise ratio (SNR). SNR is defined as
$\Var(\xb\betab)/\Var(\epsilonb)$ and we vary ${\rm SNR} \in \{15, 10,
5, 1\}$.

{\it Simulation 2:} [Changing the number of non-zero elements in
$\Bb_j$] The following scenario is used to evaluate the performance of
the methods as the number of non-zero elements in a row of $\Bb$
varies. We set $(n, p, s) = (100, 500, 10)$ and vary the number of
outputs $T \in \{500, 750, 1000\}$. For each number of outputs $T$, we
vary $T_{\rm non-zero} \in \{0.8T, 0.5T, 0.2T\}$. The samples $\xb_i$
and regression coefficients $\Bb$ are given as in Simulation 1, i.e.,
$\xb_i$ is drawn from a multivariate standard normal distribution and
the non-zero coefficients $\Bb$ are given as $\beta_{t, j} =
(-1)^{u}(4n^{-1/2}\log n + |z|)$, where $u \sim {\rm Bernoulli}(0.4)$
and $z \sim \Ncal(0, 1)$. The noise is Gaussian, with the standard
deviation defined through the SNR, which varies in $\{10, 5, 1\}$.

\begin{figure}[!h]
  \centering
  \subfigure[$\rho = 0.2$]{
    \includegraphics[width=0.45\textwidth]{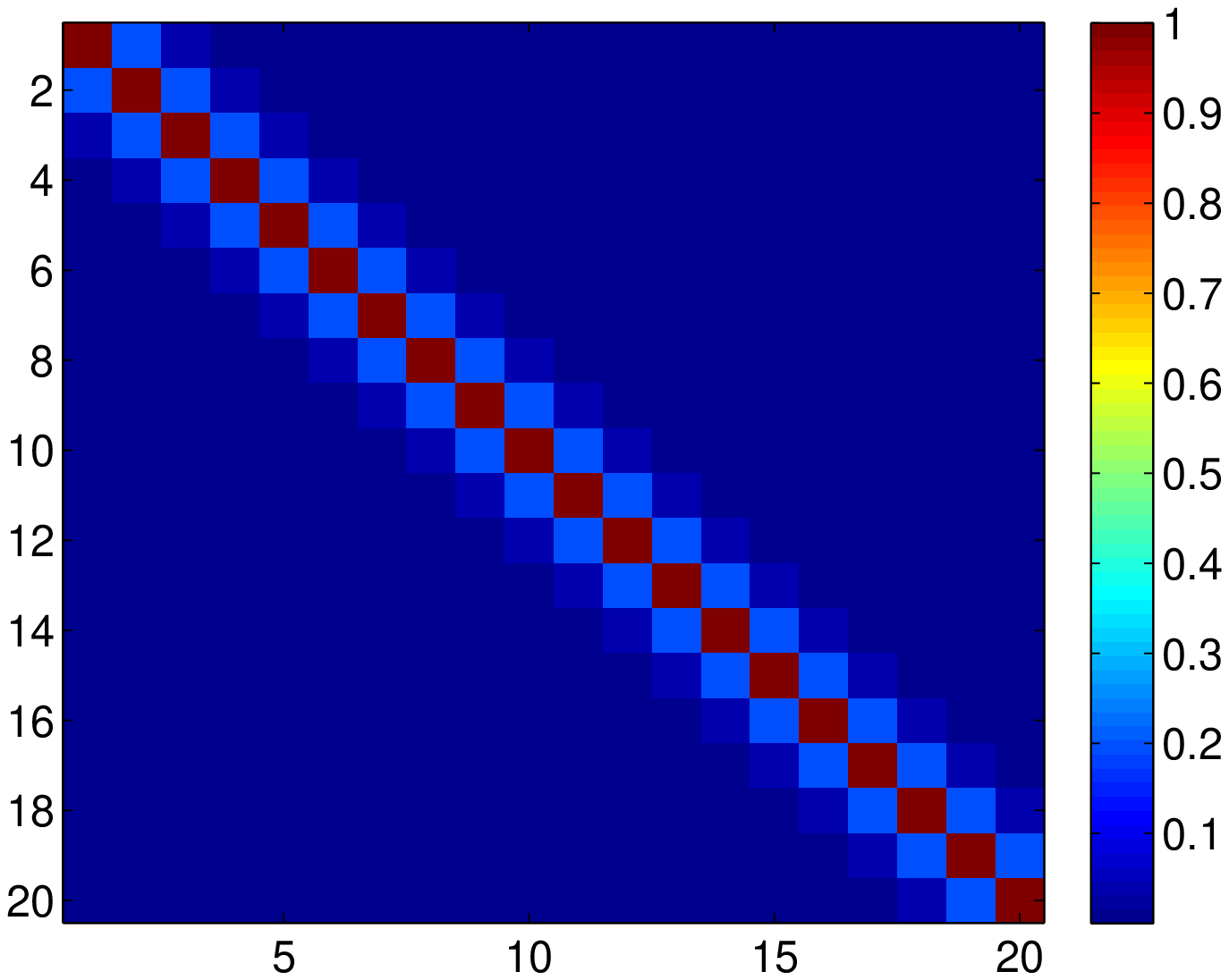}
  }
  \subfigure[$\rho = 0.7$]{
    \includegraphics[width=0.45\textwidth]{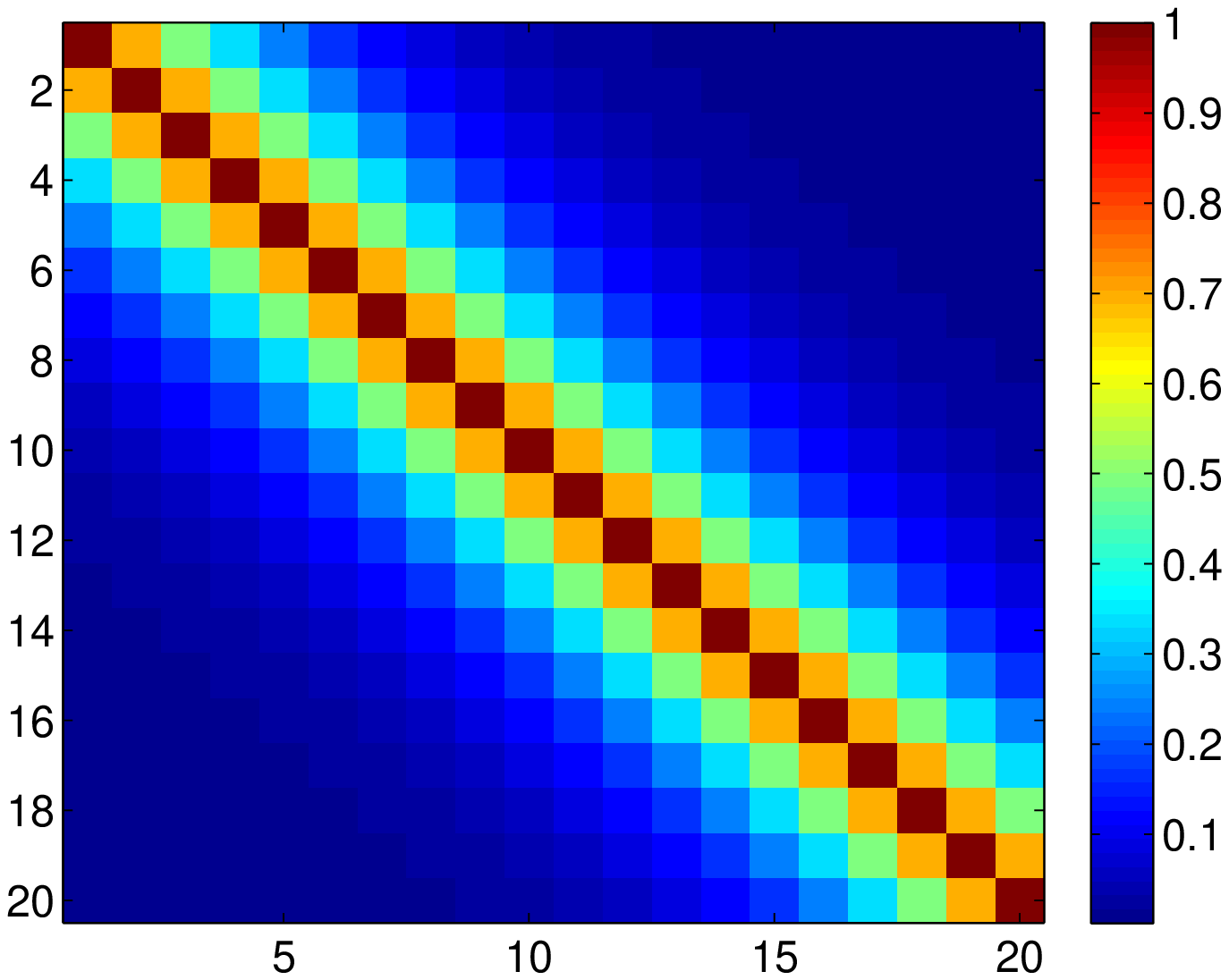}
  }
  \caption{Visualization of the correlation matrix in Simulation
    3. Only an upper left corner is presented corresponding to 20 of the 5000
    variables.}
  \label{fig:sim3-cor}
\end{figure}

{\it Simulation 3:} [Model with the decaying correlation between
variables] The following model is borrowed from
\cite{wang09forward}. We assume a correlation structure between
variables given as $\Var(\Xb_{j_1}, \Xb_{j_2}) = \rho^{|j_1 - j_2|}$,
where $\rho \in \{ 0.2, 0.5, 0.7 \}$. This correlation structure
appears naturally among ordered variables. We set $(n, p, s, T) =
(100, 5000, 3, 150)$ and $T_{\rm non-zero} = 80$. The relevant
variables are at positions $(1, 4, 7)$ and non-zero coefficients are
given as $3, 1.5$ and $2$ respectively. The SNR varies in $\{10, 5,
1\}$. A heat map of the correlation matrix between different
covariates is given in Figure~\ref{fig:sim3-cor}.

\begin{figure}[!h]
  \centering
  \subfigure[$\rho = 0.2$]{
    \includegraphics[width=0.45\textwidth]{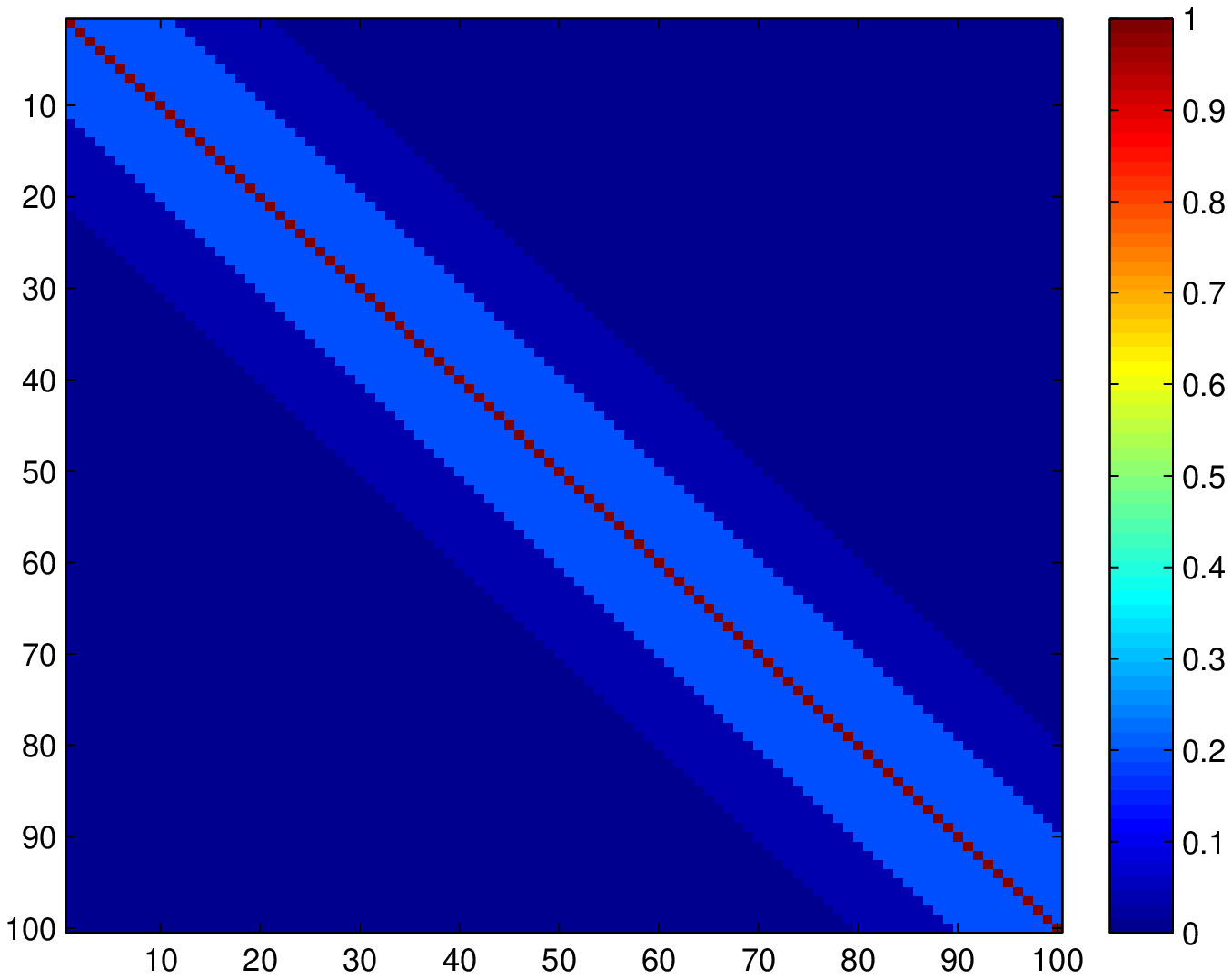}
  }
  \subfigure[$\rho = 0.5$]{
    \includegraphics[width=0.45\textwidth]{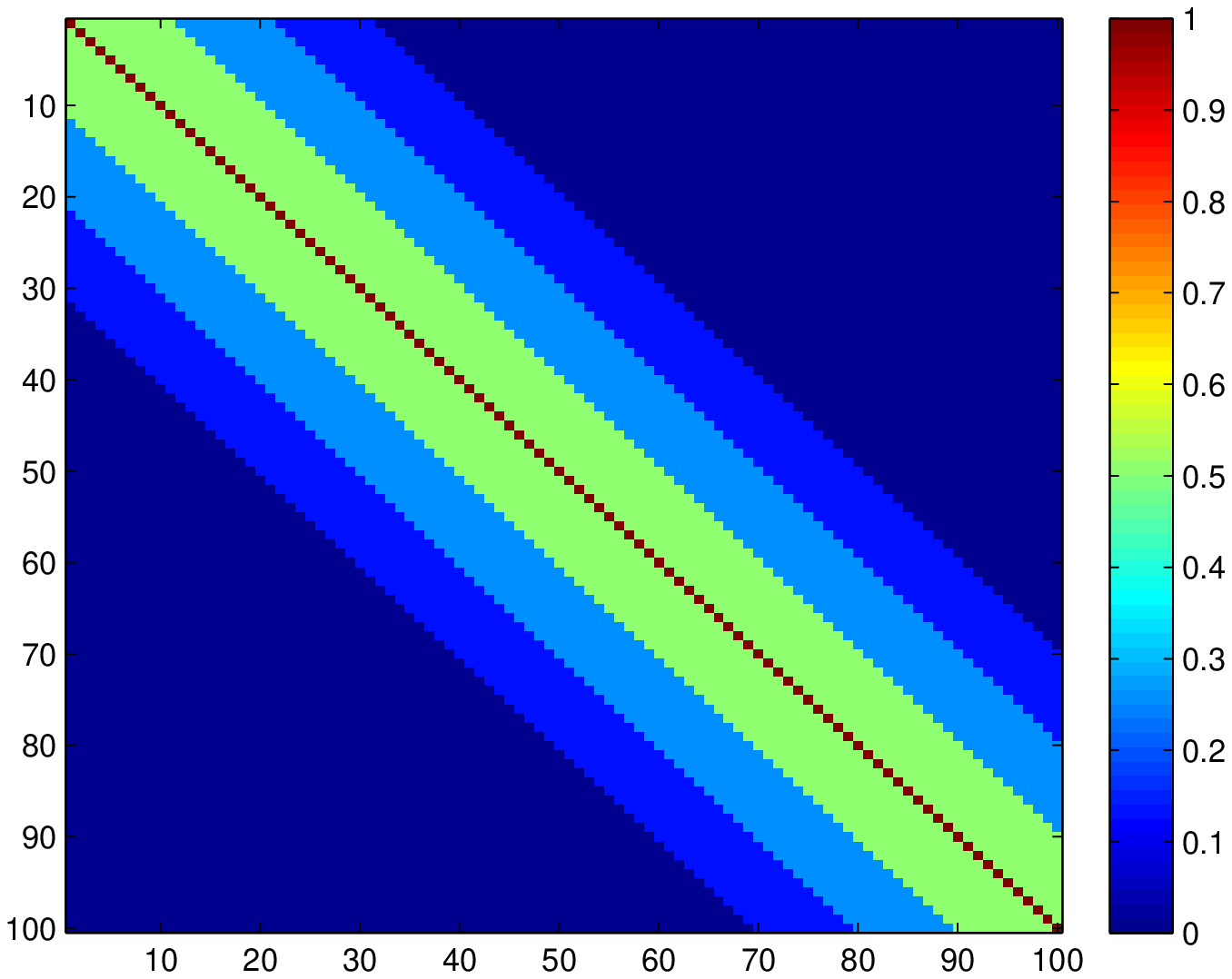}
  }
  \caption{Visualization of the correlation matrix in Simulation
    4. Only an upper left corner is presented corresponding to 100 of the 4000
    variables.}
  \label{fig:sim4}
\end{figure}

{\it Simulation 4:} [Model with the block-compound correlation
structure] The following model assumes a block compound correlation
structure. For a parameter $\rho$, the correlation between two
variables $\Xb_{j_1}$ and $\Xb_{j_2}$ is given as $\rho$, $\rho^2$ or
$\rho^3$ when $|j_1 - j_2| \leq 10$, $|j_1 - j_2| \in (10, 20]$ or
$|j_1 - j_2| \in (20, 30]$ and it is set to 0 otherwise. We set $(n,
p, s, T) = (150, 4000, 8, 150)$, $T_{\rm non-zero} = 80$ and the
parameter $\rho \in \{0.2, 0.5\}$. The relevant variables are at
located at positions 1, 11, 21, 31, 41, 51, 61, 71 and 81, so that
each block of highly correlated variables has exactly one relevant
variable. The values of relevant coefficients is given as in
Simulation 1. The noise is Gaussian and the SNR varies in $\{10, 5,
1\}$. A heat map of the correlation matrix between different
covariates is given in Figure~\ref{fig:sim4}.

{\it Simulation 5:} [Model with a 'masked' relevant variable] This
model represents a difficult setting. It is modified from
\cite{wang09forward}. We set $(n, p, s, T) = (200, 10000, 5,
500)$. The number of non-zero elements in each row varies is $T_{\rm
  non-zero} \in \{400, 250, 100\}$. For $j \in [s]$ and $t \in [T]$,
the non-zero elements equal $\beta_{t,j} = 2j$. Each row of $\Xb$ is
generated as follows. Draw independently $\zb_i$ and $\zb_i'$ from a
$p$-dimensional standard multivariate normal distribution. Now,
$x_{ij} = (z_{ij} + z'_{ij})/\sqrt(2)$ for $j \in [s]$ and $x_{ij} =
(z_{ij} + \sum_{j' \in [s]}z_{ij'})/2$ for $j \in [p]\backslash
[s]$. Now, $\Cor(x_{i,1}, y_{t,i})$ is much smaller then
$\Cor(x_{i,j}, y_{t,i})$ for $j \in [p]\backslash [s]$, so that it
becomes difficult to select variable 1. The variable 1 is 'masked'
with the noisy variables. This setting is difficult for screening
procedures as they take into consideration only marginal information.
The noise is Gaussian with standard deviation $\sigma \in \{1.5, 2.5,
4.5\}$.

In the next section we summarize results of our experimental
findings. Our simulation setting transitions from a simple scenario
considered in Simulation 1 towards a challenging one in Simulation
5. Simulation 1 is adopted from \cite{fan08sis} as a toy model on
which all algorithms should work well. Simulation 2 examines the
influence of the number of non-zero elements in a relevant row of the
matrix $\Bb$. We expect that Algorithm~\ref{alg:group_forward} will
outperform procedures that perform estimation on individual outputs
when $T_{\rm non-zero}$ is large, while when $T_{\rm non-zero}$ is
small the single-task screening procedures should have an
advantage. Our intuition is also supported by recent results of
\cite{kolar10union}.  Simulations 3 and 4 represent more challenging
situations with structured correlation that naturally appears in many
data sets, for example, a correlation between gene measurements that
are closely located on a chromosome. Finally Simulation 5 is
constructed in such a way that procedures which use only marginal
information are going to include irrelevant variables before relevant
ones. 

\subsection{Results of simulations}

\begin{table}[p]
\begin{adjustwidth}{-1in}{-1in}
  \centering \footnotesize
 \caption{
    \label{table:sim1}
    Results for simulation 1 with parameters $(n, p, s, T) =
    (500, 20000, 18, 500)$, $T_{\rm non-zero} = 500$}

  \begin{tabular}{c c c c c c c c c}    
    \hline\hline
    & 
    & Prob. (\%) of 
    & Fraction (\%) of
    & Fraction (\%) of
    & Fraction (\%) of 
    & 
    & Est. error 
    & Test error 
    \\
    & Method name 
    & $\Mcal_* \subseteq \hat S$ 
    & Correct zeros 
    & Incorrect zeros
    & $\Mcal_* = \hat S$ 
    & $|\hat S|$ 
    & $\norm{\Bb - \hat \Bb}_2^2$ 
    & $R^2$ \\
    \hline
    \multicolumn{9}{c}{SNR = 15}\\
    \\
\multirow{5}{*}{\begin{sideways}Union\end{sideways} \begin{sideways} Support\end{sideways}}
& SIS-ALASSO & 100.0 & 100.0 & 0.0 & 10.0 & 20.2 & - & -\\
& ISIS-ALASSO & 100.0 & 100.0 & 0.0 & 18.0 & 19.6 & - & -\\
& OMP & 100.0 & 100.0 & 0.0 & 0.0 & 23.9 & - & -\\
& S-OMP & 100.0 & 100.0 & 0.0 & 100.0 & 18.0 & - & -\\
& S-OMP-ALASSO & 100.0 & 100.0 & 0.0 & 100.0 & 18.0 & - & -\\
\\
\multirow{4}{*}{\begin{sideways}Exact\end{sideways} \begin{sideways} Support\end{sideways}}
& SIS-ALASSO & 0.0 & 100.0 & 0.7 & 0.0 & 8940.5 & 0.97 & 0.93 \\
& ISIS-ALASSO & 100.0 & 100.0 & 0.0 & 18.0 & 9001.6 & 0.33 & 0.93 \\
& OMP & 100.0 & 100.0 & 0.0 & 0.0 & 9005.9 & 0.20 & 0.93 \\
& S-OMP-ALASSO & 100.0 & 100.0 & 0.0 & 100.0 & 9000.0 & 0.20 & 0.93 \\
\hline

  \end{tabular}
\end{adjustwidth}
\end{table}

\begin{table}[p]
  \begin{adjustwidth}{-1in}{-1in}
    \caption{Results for simulation 2 with parameters $(n, p, s, T) =
      (200, 5000, 10, 1000)$, $T_{\rm non-zero} = 200$}
    \centering \footnotesize
  \begin{tabular}{c c c c c c c c c}
    
    \hline\hline
    & 
    & Prob. (\%) of 
    & Fraction (\%) of
    & Fraction (\%) of
    & Fraction (\%) of 
    & 
    & Est. error 
    & Test error 
    \\
    & Method name 
    & $\Mcal_* \subseteq \hat S$ 
    & Correct zeros 
    & Incorrect zeros
    & $\Mcal_* = \hat S$ 
    & $|\hat S|$ 
    & $\norm{\Bb - \hat \Bb}_2^2$ 
    & $R^2$ \\
    \hline

    \multicolumn{9}{c}{SNR = 5}\\
    \\
\multirow{5}{*}{\begin{sideways}Union\end{sideways} \begin{sideways} Support\end{sideways}}
& SIS-ALASSO & 100.0 & 100.0 & 0.0 & 100.0 & 10.0 & - & -\\
& ISIS-ALASSO & 100.0 & 100.0 & 0.0 & 100.0 & 10.0 & - & -\\
& OMP & 100.0 & 97.4 & 0.0 & 0.0 & 139.6 & - & -\\
& S-OMP & 100.0 & 100.0 & 0.0 & 100.0 & 10.0 & - & -\\
& S-OMP-ALASSO & 100.0 & 100.0 & 0.0 & 100.0 & 10.0 & - & -\\
\\
\multirow{4}{*}{\begin{sideways}Exact\end{sideways} \begin{sideways} Support\end{sideways}}
& SIS-ALASSO & 100.0 & 100.0 & 0.0 & 100.0 & 2000.0 & 0.04 & 0.72 \\
& ISIS-ALASSO & 100.0 & 100.0 & 0.0 & 100.0 & 2000.0 & 0.04 & 0.72 \\
& OMP & 100.0 & 100.0 & 0.0 & 0.0 & 2131.6 & 0.05 & 0.71 \\
& S-OMP-ALASSO & 100.0 & 100.0 & 0.0 & 100.0 & 2000.0 & 0.03 & 0.72 \\
\hline

    \label{table:sim2}

  \end{tabular}
\end{adjustwidth}

\end{table}

\begin{table}[p]
  \begin{adjustwidth}{-1in}{-1in}

  \caption{
    \label{table:sim3}
    Results for simulation 3 with parameters $(n, p, s, T) = (100,
    5000, 3,150)$, $T_{\rm non-zero} = 80$, $\rho = 0.5$ }
  
  \centering \footnotesize

  \begin{tabular}{c c c c c c c c c}
    
    \hline\hline
    & 
    & Prob. (\%) of 
    & Fraction (\%) of
    & Fraction (\%) of
    & Fraction (\%) of 
    & 
    & Est. error 
    & Test error 
    \\
    & Method name 
    & $\Mcal_* \subseteq \hat S$ 
    & Correct zeros 
    & Incorrect zeros
    & $\Mcal_* = \hat S$ 
    & $|\hat S|$ 
    & $\norm{\Bb - \hat \Bb}_2^2$ 
    & $R^2$ \\
    \hline

    \multicolumn{9}{c}{SNR = 5}\\
    \\
\multirow{5}{*}{\begin{sideways}Union\end{sideways} \begin{sideways} Support\end{sideways}}
& SIS-ALASSO & 100.0 & 100.0 & 0.0 & 97.0 & 3.0 & - & -\\
& ISIS-ALASSO & 100.0 & 100.0 & 0.0 & 96.0 & 3.0 & - & -\\
& OMP & 100.0 & 99.8 & 0.0 & 0.0 & 19.6 & - & -\\
& S-OMP & 100.0 & 100.0 & 0.0 & 100.0 & 3.0 & - & -\\
& S-OMP-ALASSO & 100.0 & 100.0 & 0.0 & 100.0 & 3.0 & - & -\\
\\
\multirow{4}{*}{\begin{sideways}Exact\end{sideways} \begin{sideways} Support\end{sideways}}
& SIS-ALASSO & 60.0 & 100.0 & 0.2 & 57.0 & 239.5 & 0.10 & 0.61 \\
& ISIS-ALASSO & 84.0 & 100.0 & 0.1 & 80.0 & 239.8 & 0.08 & 0.61 \\
& OMP & 100.0 & 100.0 & 0.0 & 0.0 & 256.6 & 0.06 & 0.61 \\
& S-OMP-ALASSO & 100.0 & 100.0 & 0.0 & 100.0 & 240.0 & 0.03 & 0.62 \\
\hline

  \end{tabular}
  \end{adjustwidth}
\end{table}

\begin{table}[tp]
  \begin{adjustwidth}{-1in}{-1in}

    \caption{
      \label{table:sim4}
      Results of simulation 4 with parameters $(n, p, s, T) = (150, 4000, 8,
      150)$, $T_{\rm non-zero} = 80$, $\rho = 0.5$}

  \centering \footnotesize

  \begin{tabular}{c c c c c c c c c}
    
    \hline\hline
    & 
    & Prob. (\%) of 
    & Fraction (\%) of
    & Fraction (\%) of
    & Fraction (\%) of 
    & 
    & Est. error 
    & Test error 
    \\
    & Method name 
    & $\Mcal_* \subseteq \hat S$ 
    & Correct zeros 
    & Incorrect zeros
    & $\Mcal_* = \hat S$ 
    & $|\hat S|$ 
    & $\norm{\Bb - \hat \Bb}_2^2$ 
    & $R^2$ \\
    \hline

    \multicolumn{9}{c}{SNR = 10}\\
    \\
\multirow{5}{*}{\begin{sideways}Union\end{sideways} \begin{sideways} Support\end{sideways}}
& SIS-ALASSO & 100.0 & 100.0 & 0.0 & 100.0 & 8.0 & - & -\\
& ISIS-ALASSO & 100.0 & 100.0 & 0.0 & 97.0 & 8.0 & - & -\\
& OMP & 100.0 & 99.9 & 0.0 & 2.0 & 11.7 & - & -\\
& S-OMP & 100.0 & 100.0 & 0.0 & 100.0 & 8.0 & - & -\\
& S-OMP-ALASSO & 100.0 & 100.0 & 0.0 & 100.0 & 8.0 & - & -\\
\\
\multirow{4}{*}{\begin{sideways}Exact\end{sideways} \begin{sideways} Support\end{sideways}}
& SIS-ALASSO & 35.0 & 100.0 & 1.4 & 35.0 & 631.3 & 0.55 & 0.88 \\
& ISIS-ALASSO & 100.0 & 100.0 & 0.0 & 97.0 & 640.0 & 0.14 & 0.89 \\
& OMP & 100.0 & 100.0 & 0.0 & 2.0 & 643.7 & 0.10 & 0.89 \\
& S-OMP-ALASSO & 100.0 & 100.0 & 0.0 & 100.0 & 640.0 & 0.09 & 0.89 \\
\hline

  \end{tabular}
  \end{adjustwidth}
\end{table}

\begin{table}[tp]
  \begin{adjustwidth}{-1in}{-1in}

    \caption{
      \label{table:sim5}
      Results of simulation 5 with parameters $(n, p, s, T) = (200, 10000, 5,
      500)$, $T_{\rm non-zero} = 400$ }

  \centering \footnotesize

  \begin{tabular}{c c c c c c c c c}
    
    \hline\hline
    & 
    & Prob. (\%) of 
    & Fraction (\%) of
    & Fraction (\%) of
    & Fraction (\%) of 
    & 
    & Est. error 
    & Test error 
    \\
    & Method name 
    & $\Mcal_* \subseteq \hat S$ 
    & Correct zeros 
    & Incorrect zeros
    & $\Mcal_* = \hat S$ 
    & $|\hat S|$ 
    & $\norm{\Bb - \hat \Bb}_2^2$ 
    & $R^2$ \\
    \hline

    \multicolumn{9}{c}{$\sigma$ = 1.5}\\
    \\
\multirow{5}{*}{\begin{sideways}Union\end{sideways} \begin{sideways} Support\end{sideways}}
& SIS-ALASSO & 53.0 & 99.6 & 9.4 & 0.0 & 41.1 & - & -\\
& ISIS-ALASSO & 100.0 & 99.8 & 0.0 & 0.0 & 28.1 & - & -\\
& OMP & 100.0 & 99.9 & 0.0 & 12.0 & 10.0 & - & -\\
& S-OMP & 100.0 & 100.0 & 0.0 & 44.0 & 5.6 & - & -\\
& S-OMP-ALASSO & 100.0 & 100.0 & 0.0 & 100.0 & 5.0 & - & -\\
\\
\multirow{4}{*}{\begin{sideways}Exact\end{sideways} \begin{sideways} Support\end{sideways}}
& SIS-ALASSO & 0.0 & 100.0 & 68.9 & 0.0 & 936.0 & 84.66 & 0.66 \\
& ISIS-ALASSO & 0.0 & 100.0 & 16.2 & 0.0 & 1791.9 & 5.80 & 0.96 \\
& OMP & 100.0 & 100.0 & 0.0 & 12.0 & 2090.3 & 0.06 & 0.99 \\
& S-OMP-ALASSO & 100.0 & 100.0 & 0.0 & 100.0 & 2000.0 & 0.05 & 0.99 \\
\hline

  \end{tabular}
  \end{adjustwidth}
\end{table}

Tables giving detailed results of the above described simulations are
given in the Appendix. In this section, we outline main findings and
reproduce some parts of the tables that we think are insightful.

Table~\ref{table:sim1} shows parts of the results for simulation 1. We
can see that all methods perform well in the setting when the input
variables are mutually uncorrelated and the SNR is high. Note that
even though the variables are uncorrelated, the sample correlation
between variables can be quite high due to large $p$ and small $n$,
which can result in selection of spurious variables. As we can see
from the table, comparing to SIS, ISIS and OMP, the S-OMP is able to
select the correct union support, while the procedures that select
variables based on different outputs separately also include
additional spurious variables into the selection. Furthermore, we can
see that the S-OMP-ALasso procedure does much better on the problem of
exact support recovery compared to the other procedures.  The first
simulations suggests that somewhat higher computational cost of the
S-OMP procedure can be justified by the improved performance on the
problem of union and exact support recovery as well as on the error in
the estimated coefficients.

Table~\ref{table:sim2} shows parts of the results for simulation 2. In
this simulation we measured the performance of estimation procedures
as the amount of shared input variables between different outputs
varies. The parameter $T_{\rm non-zero}$ controls the amount of
information that is shared between different tasks as defined in the
previous subsection. In particular, the parameter controls the number
of non-zero elements in a row of the matrix $\Bb$ corresponding to a
relevant variable. When the number of non-zero elements is high, a
variable is relevant to many tasks and we say that outputs overlap. In
this setting the S-OMP procedure is expected to outperform the other
methods, however, when $T_{\rm non-zero}$ is low, the noise coming
from the tasks for which the variable is irrelevant can actually harm
the performance.  The table shows results when the overlap of shared
variables is small, that is, a relevant variable is only relevant for
10\% of outputs. As one could expect, the S-OMP procedure does as well
as other procedures. This is not surprising, since the amount of
shared information between different outputs is limited. Therefore, if
one expects little variable sharing across different outputs, using
the SIS or ISIS may results in similar accuracy, but an improved
computational efficiency. It is worth pointing out that in our
simulations, the different tasks are correlated since the same design
$\Xb$ used for all tasks. However, we expect the same qualitative
results even under the model given in equation \eqref{eq:model_linear}
where different tasks can have different designs $\Xb_t$ and the
outputs are uncorrelated.

Simulation 3 represents a situation that commonly occur in nature,
where there is an ordering among input variables and the correlation
between variables decays as the distance between variables
increase. The model in simulation 4 is a modification of the model in
simulation 3 where the variables are grouped and there is some
correlation between different groups. Table~\ref{table:sim3} gives
results for simulation 3 for the parameter $\rho=0.5$. In this
setting, the S-OMP performs much better that the other procedures. The
improvement becomes more pronounced with increase of the correlation
parameter $\rho$. Similar behavior is observed in simulation 4 as
well, see table~\ref{table:sim4}. Results of simulation 5, given in
Table~\ref{table:sim5}, further reinforce our intuition that the S-OMP
procedure does well even on problems with high-correlation between the
set of relevant input variables and the set of irrelevant ones.

To further compare the performance of the S-OMP procedure to the SIS,
we explore the minimum number of iterations needed for the algorithm
to include all the relevant variables into the selected model. From
our limited numerical experience, we note that the simulation
parameters do not affect the number of iterations for the S-OMP
procedure. This is unlike the SIS procedure, which occasionally
requires a large number of steps before all the true variables are
included, see Figure 3 in \cite{fan08sis}. We note that while the
S-OMP procedure does include, in many cases, all the relevant
variables before the irrelevant ones, the BIC criterion is not able to
correctly select the number of variables to include, when the SNR is
small. As a result, we can see the drop in performance as the SNR
decreases.

\subsection{Real data analysis}

We demonstrate an application of the S-OMP to a genome-wide
association mapping problem. The data were collected by our
collaborator Judie Howrylak, M.D. at Harvard Medical
School from 200 individuals that are suffering
from asthma. For each individual, we have a collection of about $\sim
350 000$ genetic markers\footnote{These markers were preprocessed, by
  imputing missing values and removing duplicate SNPs that were
  perfectly correlated with other SNPs.}, which are called single
nucleotide polymorphisms (SNPs), and a collection of 1424 gene
expression measurements. The goal of this study is to identify a small
number of SNPs that can help explain variations in gene
expressions. Typically, this type of analysis is done by regressing
each gene individually on the measured SNPs, however, since the data
are very noisy, such an approach results selecting many variables. Our
approach to this problem here is to regress a group of genes onto the
SNPs instead.  There has been some previous work on this problem
\cite{kim09statistical}, that considered regressing groups of genes
onto SNPs, however, those approaches use variants of the estimation
procedure given in Eq.~\eqref{eq:group_optimization}, which is not
easily scalable to the data we analyze here.

We use the spectral relaxation of the k-means clustering
\cite{zha01spectral} to group 1424 genes into 48 clusters according to
their expression values, so that the minimum, maximum and median
number of genes per cluster is 4, 90 and 19, respectively. The number
of clusters was chosen somewhat arbitrarily, based on the domain
knowledge of the medical experts. The main idea behind the clustering
is that we want to identify genes that belong to the same regulatory
pathway since they are more likely to be affected with the same
SNPs. Instead of clustering, one may use prior knowledge to identify
interesting groups of genes. Next, we want to use the S-OMP procedure
to identify relevant SNPs for each of the gene clusters.  Since, we do
not have the ground truth for the data set, we use predictive power on
the test set and the size of estimated models to access their
quality. We randomly split the data into a training set of size 170
and a testing set of size 30 and report results over 500 runs. We
compute the $R^2$ coefficient on the test set defined as
$1-30^{-1}T^{-1}\sum_{t \in [T]}\norm{\yb_{t, {\rm test}} - \Xb_{t,
    {\rm test}}\hat \betab_t}_2^2$ (because the data has been
normalized).

\begin{table}[!hp]
  \caption{Results on the asthma data}
  \footnotesize
  \begin{tabular}{c c c c}
    
    \hline\hline
    & Method name 
    & Union support
    & $R^2$ \\
    \hline

    \multirow{3}{1in}{Cluster  9\\ Size = 18} 
& SIS-ALASSO & 18.0 (1.0) & 0.178 (0.006) \\ 
& OMP & 17.5 (2.9) & 0.167 (0.002) \\ 
& S-OMP & 1.0 (0.0) & 0.214 (0.005) \\ 
 \\\multirow{3}{1in}{Cluster 16\\ Size = 31} 
& SIS-ALASSO & 31.0 (1.0) & 0.160 (0.007) \\ 
& OMP & 29.0 (1.8) & 0.165 (0.002) \\ 
& S-OMP & 1.0 (0.0) & 0.209 (0.005) \\ 
 \\\multirow{3}{1in}{Cluster 17\\ Size = 19} 
& SIS-ALASSO & 18.5 (0.9) & 0.173 (0.006) \\ 
& OMP & 19.5 (0.8) & 0.146 (0.003) \\ 
& S-OMP & 1.0 (0.0) & 0.184 (0.004) \\ 
 \\\multirow{3}{1in}{Cluster 19\\ Size = 17} 
& SIS-ALASSO & 17.0 (1.2) & 0.270 (0.017) \\ 
& OMP & 11.0 (4.1) & 0.213 (0.008) \\ 
& S-OMP & 1.0 (0.0) & 0.280 (0.017) \\ 
 \\\multirow{3}{1in}{Cluster 22\\ Size = 34} 
& SIS-ALASSO & 34.0 (0.9) & 0.153 (0.005) \\ 
& OMP & 30.0 (7.3) & 0.142 (0.000) \\ 
& S-OMP & 1.0 (0.0) & 0.145 (0.002) \\ 
 \\\multirow{3}{1in}{Cluster 23\\ Size = 35} 
& SIS-ALASSO & 35.0 (0.9) & 0.238 (0.018) \\ 
& OMP & 33.0 (9.9) & 0.208 (0.009) \\ 
& S-OMP & 1.0 (0.0) & 0.229 (0.014) \\ 
 \\\multirow{3}{1in}{Cluster 24\\ Size = 28} 
& SIS-ALASSO & 28.0 (1.0) & 0.123 (0.003) \\ 
& OMP & 28.0 (2.6) & 0.114 (0.001) \\ 
& S-OMP & 1.0 (0.0) & 0.129 (0.003) \\ 
 \\\multirow{3}{1in}{Cluster 32\\ Size = 15} 
& SIS-ALASSO & 15.0 (0.9) & 0.188 (0.010) \\ 
& OMP & 10.0 (2.6) & 0.211 (0.006) \\ 
& S-OMP & 1.0 (0.0) & 0.215 (0.008) \\ 
 \\\multirow{3}{1in}{Cluster 36\\ Size = 33} 
& SIS-ALASSO & 34.0 (1.4) & 0.147 (0.005) \\ 
& OMP & 29.0 (5.3) & 0.157 (0.002) \\ 
& S-OMP & 1.0 (0.0) & 0.168 (0.004) \\ 
 \\\multirow{3}{1in}{Cluster 37\\ Size = 19} 
& SIS-ALASSO & 19.0 (0.9) & 0.207 (0.015) \\ 
& OMP & 22.0 (2.5) & 0.175 (0.006) \\ 
& S-OMP & 1.0 (0.0) & 0.235 (0.014) \\ 
 \\\multirow{3}{1in}{Cluster 39\\ Size = 24} 
& SIS-ALASSO & 24.0 (0.9) & 0.131 (0.006) \\ 
& OMP & 27.0 (1.9) & 0.141 (0.003) \\ 
& S-OMP & 1.0 (0.0) & 0.160 (0.005) \\ 
 \\\multirow{3}{1in}{Cluster 44\\ Size = 35} 
& SIS-ALASSO & 35.0 (0.9) & 0.177 (0.010) \\ 
& OMP & 26.5 (6.6) & 0.183 (0.005) \\ 
& S-OMP & 1.0 (0.0) & 0.170 (0.011) \\ 
 \\\multirow{3}{1in}{Cluster 49\\ Size = 23} 
& SIS-ALASSO & 23.0 (1.0) & 0.124 (0.004) \\ 
& OMP & 23.0 (1.2) & 0.140 (0.000) \\ 
& S-OMP & 1.0 (0.0) & 0.159 (0.004) \\ 
 \\
    \hline
    \label{table:real-small}
  \end{tabular}
\end{table}

Due to space constraints, we give results on few clusters in
Table~\ref{table:real-small} and note that, qualitatively, the results
do not vary much between different clusters.  While the fitted models
have limited predictive performance, which results from highly noisy
data, we observe that the S-OMP is able to identify on average one SNP
per gene cluster that is related to a large number of genes. Other
methods, while having a similar predictive performance, select a
larger number of SNPs which can be seen from the size of the union
support. On this particular data set, the S-OMP seems produce results
that are more interpretable from a specialist's points of
view. Further investigation needs to be done to verify the biological
significance of the selected SNPs, however, the details of such an
analysis are going to be reported elsewhere.

\section{Conclusions}

In this work, we analyze the Simultaneous Orthogonal Matching Pursuit
as a method for variable selection in an ultra-high dimensional
space. We prove that the S-OMP is screening consistent and provide a
practical way to select the number of steps in the procedure using the
modified Bayesian information criterion. Our limited numerical
experience shows that the method performs well in practice and that
the joint estimation from multiple outputs often outperforms methods
that use one regression output at the time. Furthermore, we can see
the S-OMP procedure as way to improve the variable selection
properties of the SIS without having to solve a costly complex
optimization procedure in Eq.~\eqref{eq:group_optimization},
therefore, balancing the computational costs and the estimation
accuracy.

\section{Appendix}

\subsection{Proof of Theorem~\ref{thm:path_screening_consistent}}

Under the assumptions of the theorem, the number of relevant variables
$s$ is relatively small compared to the sample size $n$. The proof
strategy can be outlined as follows: i) we are going to show that,
with high probability, at least one relevant variable is going to be
identified within the following $m_{\rm one}^*$ steps, conditioning on
the already selected variables $\Mcal^{(k)}$ and this holds uniformly
for all $k$; ii) we can conclude that all the relevant variables are
going to be selected within $m_{\max}^* = sm_{\rm one}^*$ steps. Exact
values for $m_{\rm one}^*$ and $m_{\max}^*$ are given below. Without
loss of generality, we analyze the first step of the algorithm, i.e.,
we show that the first relevant variable is going to be selected
within the first $m_{\rm one}^*$ steps.

Assume that in the first $m_{\rm one}^*-1$ steps, there were no
relevant variables selected. Assuming that the variable selected in
the $m_{\rm one}^*$-th step is still an irrelevant one, we will arrive
at a contradiction, which shows that at least one relevant variable
has been selected in the first $m_{\rm one}^*$ steps. For any step
$k$, the reduction of the squared error is given as
\begin{equation} \label{eq:def_delta} \Delta(k) := {\rm RSS}(k-1) -
  {\rm RSS}(k) = \sum_{t}\norm{\Hb_{t, \hat f_k}^{(k)}(\Ib_{n \times
      n} - \Hb_{t, \Mcal^{(k)}})\yb_t}_2^2
\end{equation}
with $\Hb_{t, j}^{(k)} = \Xb_{t,j}^{(k)}\Xb_{t,j}^{(k)'}
\norm{\Xb_{t,j}^{(k)}}^{-2}$ and $\Xb_{t,j}^{(k)} = (\Ib_{n \times n}
- \Hb_{t, \Mcal^{(k)}})\Xb_{t,j}$. We are interested in the quantity
$\sum_{k=1}^{m_{\rm one}^*} \Delta(k)$, when all the selected
variables $\hat f_k$ (see Algorithm~\ref{alg:group_forward}) belong to
$[p] \backslash \Mcal_*$.

In what follows, we will derive a lower bound for $\Delta(k)$. We
perform our analysis on the event
\begin{equation}
  \begin{aligned}
    \Ecal & = \{ \min_{t \in [T]} \min_{\Mcal \subseteq [p], |\Mcal| \leq
      m_{\max}^*} \Lambda_{\min} (\hat \Sigmab_{\Mcal}) \geq
      \phi_{\min}/2\} \\
      & \quad \bigcap \{
      \max_{t \in [T]} \max_{\Mcal \subseteq [p], |\Mcal| \leq
          m_{\max}^*} \Lambda_{\max} (\hat \Sigmab_{\Mcal}) \leq
      2\phi_{\max} \}.
    \end{aligned}
\end{equation}
From the definition of $\hat f_k$, we have
\begin{equation} \label{eq:lower_bound_on_delta}
  \begin{aligned}
    \Delta(k) & \geq \max_{j \in \Mcal_*} 
    \sum_{t}\norm{\Hb_{t, j}^{(k)} 
      (\Ib_{n \times n} - \Hb_{t, \Mcal^{(k)}})\yb_t}_2^2 \\
    & \geq \max_{j \in \Mcal_*} \ \Big(\sum_{t}\norm{\Hb_{t, j}^{(k)}
       (\Ib_{n \times n} - \Hb_{t, \Mcal^{(k)}}) \Xb_{t, \Mcal_*}
        \betab_{t, \Mcal_*}}_2^2 \\
    & \qquad - \sum_{t} \norm{\Hb_{t, j}^{(k)}
       (\Ib_{n \times n} - \Hb_{t, \Mcal^{(k)}})\epsilonb_t}_2^2
       \Big) \\
    & \geq \max_{j \in \Mcal_*}\ \sum_{t}\norm{\Hb_{t, j}^{(k)}
       (\Ib_{n \times n} - \Hb_{t, \Mcal^{(k)}}) \Xb_{t, \Mcal_*}
        \betab_{t, \Mcal_*}}_2^2 \\
    & \qquad - \max_{j \in \Mcal_*}\ 
       \sum_{t} \norm{\Hb_{t, j}^{(k)}
       (\Ib_{n \times n} - \Hb_{t, \Mcal^{(k)}})\epsilonb_t}_2^2 \\
    & = (I) - (II).
  \end{aligned}
\end{equation}
We deal with these two terms separately. Let $\Hb_{t, \Mcal}^{\perp} =
\Ib_{n \times n} - \Hb_{t, \Mcal}$ denote the projection matrix. We
have that the first term $(I)$ is lower bounded by
\begin{equation} \label{eq:term_i_1}
  \begin{aligned}
    \max_{j \in \Mcal_*}&\ \sum_{t}\norm{\Hb_{t, j}^{(k)}
      \Hb_{t, \Mcal^{(k)}}^{\perp} \Xb_{t, \Mcal_*}
      \betab_{t, \Mcal_*}}_2^2 \\
    & = \max_{j \in \Mcal_*}\ \sum_{t}\norm{\Xb_{t, j}^{(k)}}_2^{-2}
      |\Xb_{t, j}^{(k)'}\Hb_{t, \Mcal^{(k)}}^{\perp} \Xb_{t, \Mcal_*}
      \betab_{t, \Mcal_*}|^2 \\
    & \geq \min_{t \in [T], j \in \Mcal_*} \{ \norm{\Xb_{t, j}^{(k)}}_2^{-2} \} 
      \max_{j \in \Mcal_*} \sum_{t}
      |\Xb_{t, j}^{(k)'}\Hb_{t, \Mcal^{(k)}}^{\perp} \Xb_{t, \Mcal_*}
      \betab_{t, \Mcal_*}|^2 \\
    & \geq \{ \max_{t \in [T], j \in \Mcal_*} 
           \norm{\Xb_{t,j}}_2^2 \}^{-1}
      \max_{j \in \Mcal_*} \sum_{t}
      |\Xb_{t, j}'\Hb_{t, \Mcal^{(k)}}^{\perp} \Xb_{t, \Mcal_*}
      \betab_{t, \Mcal_*}|^2,
  \end{aligned}
\end{equation}
where the last inequality follows from the fact that
$\norm{\Xb_{t,j}}_2 \geq \norm{\Xb_{t, j}^{(k)}}_2$ and
$\Xb_{t,j}^{(k)'}\Hb_{t, \Mcal^{(k)}}^{\perp} =
\Xb_{t,j}'\Hb_{t,\Mcal^{(k)}}^{\perp}$. A simple calculation shows
that
\begin{equation} \label{eq:term_i_2}
  \begin{aligned}
    \sum_{t}&\norm{\Hb_{t,\Mcal^{(k)}}^{\perp}\Xb_{t, \Mcal_*}
      \betab_{t, \Mcal_*}}_2^2 \\
    & = \sum_{t} \sum_{j \in \Mcal_*} \beta_{t,j} \Xb_{t, j}
      \Hb_{t,\Mcal^{(k)}}^{\perp}\Xb_{t, \Mcal_*} \betab_{t, \Mcal_*} \\
    & \leq  \sum_{j \in \Mcal_*} \sqrt{\sum_{t} \beta_{t,j}^2}
      \sqrt{\sum_{t} (\Xb_{t, j} \Hb_{t,\Mcal^{(k)}}^{\perp}
              \Xb_{t, \Mcal_*} \betab_{t, \Mcal_*})^2} \\
    & \leq \norm{\betab}_{2,1} \max_{j \in \Mcal_*} 
    \sqrt{\sum_{t} (\Xb_{t, j} \Hb_{t,\Mcal^{(k)}}^{\perp}
              \Xb_{t, \Mcal_*} \betab_{t, \Mcal_*})^2}.
  \end{aligned}
\end{equation}
Plugging \eqref{eq:term_i_2} back into \eqref{eq:term_i_1}, the
following lower bound is achieved
\begin{equation} \label{eq:term_I_3} 
  (I) \geq \{ \max_{t \in [T], j \in \Mcal_*} 
  \norm{\Xb_{t,j}}_2^2 \}^{-1} \frac{ (\sum_{t}
    \norm{\Hb_{t,\Mcal^{(k)}}^{\perp} \Xb_{t, \Mcal_*}
      \betab_{t, \Mcal_*}}_2^2)^2 }{ \norm{\Bb}_{2,1}^2 }.
\end{equation}
On the event $\Ecal$, $\max_{t \in [T], j \in \Mcal_*}
\norm{\Xb_{t,j}}_2^2 \leq 2n\phi_{\max}$. Since we have assumed that no
additional relevant predictors have been selected by the procedure, it
holds that $\Mcal_* \not\subseteq \Mcal^{(k)}$. This leads to
\begin{equation}
\sum_{t}\norm{\Hb_{t,\Mcal^{(k)}}^{\perp} \Xb_{t, \Mcal_*} \betab_{t,
    \Mcal_*}}_2^2 
\geq 2^{-1}n\phi_{\min} \min_{j \in \Mcal^*} \sum_{t \in [T]} \beta^2_{t,j},
\end{equation}
on the event $\Ecal$. Using the Cauchy–-Schwarz inequality,
$\norm{\Bb}_{2,1}^{-2} \geq s^{-1}T^{-1}C_{\betab}^{-2}$. Plugging
back into \eqref{eq:term_I_3}, we have that
\begin{equation}
  \begin{aligned}
    (I) & \geq 2^{-3}\phi_{\min}^2\phi_{\max}^{-1}C_{\betab}^{-2} 
    ns^{-1}T^{-1}(\min_{j \in \Mcal^*} \sum_{t \in [T]} \beta^2_{t,j})^2 \\
    & \geq 2^{-3}\phi_{\min}^2\phi_{\max}^{-1}C_{\betab}^{-2} 
      C_s^{-1}n^{1 - \delta_s}
      T^{-1}(\min_{j \in \Mcal^*} \sum_{t \in [T]} \beta^2_{t,j})^2
  \end{aligned}
\end{equation}

Next, we deal with the second term in
\eqref{eq:lower_bound_on_delta}. Recall that $\Xb_{t,j}^{(k)} =
\Hb_{t, \Mcal^{(k)}}^{\perp}\Xb_{t,j}$, so that
$\norm{\Xb_{t,j}^{(k)}}_2^2 \geq 2^{-1}n\phi_{\min}$, on the event
$\Ecal$.  We have
\begin{equation} \label{eq:bound_II}
  \begin{aligned}
    \sum_{t} & \norm{\Hb_{t, j}^{(k)}
      (\Ib_{n \times n} - \Hb_{t, \Mcal^{(k)}})\epsilonb_t}_2^2 \\
    & = \sum_{t} \norm{\Xb_{t,j}^{(k)}}^{-2}(\Xb_{t,j}'\Hb_{t,
      \Mcal^{(k)}}^{\perp}\epsilonb_t)^2 \\
    & \leq 2\phi_{\min}^{-1}n^{-1} \max_{j \in \Mcal_*} 
           \max_{|\Mcal| \leq m_{\max}^*} \sum_{t} 
        (\Xb_{t,j}'\Hb_{t, \Mcal}^{\perp}\epsilonb_t)^2.
  \end{aligned}
\end{equation}
Under the conditions of the theorem, $\Xb_{t,j}'\Hb_{t,
  \Mcal}^{\perp}\epsilonb_t$ is normally distributed with mean 0 and
variance $\norm{\Hb_{t,\Mcal}^{\perp} \Xb_{t,j}}_2^2$. Furthermore,
\begin{equation}
  \max_{j \in \Mcal_*} \max_{|\Mcal| \leq m_{\max}^*} \max_{t \in [T]}
  \norm{\Hb_{t,\Mcal}^{\perp} \Xb_{t,j}}_2^2 \leq
  2n\phi_{\max}.
\end{equation}
Plugging back in \eqref{eq:bound_II}, we have
\begin{equation}
  (II) \leq 2^2\phi_{\min}^{-1}\phi_{\max} \max_{j \in \Mcal_*}
  \max_{|\Mcal| \leq m_{\max}^*} \chi_{T}^2,
\end{equation}
where $\chi_{T}^2$ denotes a chi-squared random variable with $T$
degrees of freedom. The total number of possibilities for $j \in
\Mcal_*$ and $|\Mcal| \leq m_{\max}^*$ is bounded by $p^{m_{\max}^* +
  2}$. Using Lemma~\ref{lem:max_chi_squared}, with
$\epsilon=T(m_{\max}^* + 2) \log p$ and applying the union bound, we
obtain
\begin{equation} \label{eq:probability_bound_II}
  \begin{aligned}
    (II) & \leq 2^3\phi_{\min}^{-1}\phi_{\max}T(m_{\max}^* + 2)\log p \\
    & \leq 9\phi_{\min}^{-1}\phi_{\max}C_pn^{\delta_p}Tm_{\max}^*
  \end{aligned}
\end{equation}
with probability at least 
\begin{equation}
1 - p^{m_{\max}^* + 2}\exp\left(-2T(m_{\max}^* + 2) \log(p)\left(1 -
2\sqrt{\frac{1}{2(m_{\max}^* + 2) \log(p)}}\right)\right).
\end{equation}

Going back to \eqref{eq:lower_bound_on_delta}, we have the
following 
\begin{equation} \label{eq:lower_bound_on_delta_final}
  \begin{aligned}
    n^{-1}T^{-1}\Delta(k) &
      \geq 2^{-3}\phi_{\min}^2\phi_{\max}^{-1}C_{\betab}^{-2} 
      C_s^{-1}n^{- \delta_s}T^{-2}
      (\min_{j \in \Mcal^*} \sum_{t \in [T]} \beta^2_{t,j})^2 \\
      & \quad -
      9\phi_{\min}^{-1}\phi_{\max}C_pn^{\delta_p-1}m_{\max}^* \\
      & \geq 2^{-3}\phi_{\min}^2\phi_{\max}^{-1}C_{\betab}^{-2} 
      C_s^{-1}c_{\betab}^2 n^{-\delta_s - 2\delta_{\min}}\\
      & \quad -
      9\phi_{\min}^{-1}\phi_{\max}C_pn^{\delta_p-1}m_{\max}^* \\
      & \geq 2^{-3}\phi_{\min}^2\phi_{\max}^{-1}C_{\betab}^{-2} 
      C_s^{-1}c_{\betab}^2 n^{-\delta_s - 2\delta_{\min}}\\
      & \quad \times
      (1 - 72\phi_{\min}^{-3}\phi_{\max}^2C_{\betab}^{2}C_p
      C_sc_{\betab}^{-2}
      n^{\delta_s + 2\delta_{\min} + \delta_p - 1}m_{\max}^*).
  \end{aligned}
\end{equation}
Since the bound in \eqref{eq:lower_bound_on_delta_final} holds
uniformly for $k \in \{1, \ldots, m_{\rm one}^*\}$, we have that
$n^{-1}T^{-1}\sum_{t \in [T]}\norm{\yb_t}_2^2 \geq
n^{-1}T^{-1}\sum_{k=1}^{m_{\rm one}^*} \Delta(k)$.  Setting 
\begin{equation}
  m_{\rm one}^* = \lfloor 2^{4}\phi_{\min}^{-2}\phi_{\max}C_{\betab}^{2}
  C_sc_{\betab}^{-2} n^{\delta_s + 2\delta_{\min}} \rfloor 
\end{equation}
and recalling that $m_{\max}^* = sm_{\rm one}^*$, the lower bound
becomes
\begin{equation} \label{eq:contradiction} 
n^{-1}T^{-1}\sum_{t \in [T]}\norm{\yb_t}_2^2 \geq
2(1 - C n^{3\delta_s + 4\delta_{\min} + \delta_p - 1}),
\end{equation}
for a positive constant $C$ independent of $p, n, s$ and $T$. Under
the conditions of the theorem, the right side of
\eqref{eq:contradiction} is bounded below by 2. We have arrived at a 
contradiction, since under the assumptions $\Var(y_{t, i}) = 1$ and by
the weak law of large numbers, $n^{-1}T^{-1}\sum_{t \in
  [T]}\norm{\yb_t}_2^2 \rightarrow 1$ in probability. Therefore, at
least one relevant variable will be selected in $m_{\rm one}^*$
steps. 

To complete the proof, we lower bound the probability in
\eqref{eq:probability_bound_II} and the probability of the event
$\Ecal$. Plugging in the value for $m_{\max}^*$, the probability in
\eqref{eq:probability_bound_II} can be lower bounded by $1 -
\exp(-C(2T - 1)n^{2\delta_s + 2\delta_{\min} + \delta_p})$ for some
positive constant $C$. The probability of the event $\Ecal$ is lower
bounded, using Lemma~\ref{lem:sparse_eigenvalue} together with the
union bound, as $1 - C_1\exp(-C_2
\frac{n^{1-6\delta_s-6\delta_{\min}}} {\max\{\log p , \log T \}})$, for
some positive constants $C_1$ and $C_2$. Both of these probabilities
converge to $1$ under the conditions of the theorem.

\subsection{Proof of Theorem~\ref{thm:bic_screening_consistent}}

To prove the theorem, we use the same strategy as in
\cite{wang09forward}. From
Theorem~\ref{thm:path_screening_consistent}, we have that $\PP[
\exists k \in \{0, \ldots, n-1\} : \Mcal_* \subseteq \Mcal^{(k)}]
\rightarrow 1$, so $k_{\min} := \min_{k \in \{0, \ldots, n-1\}}\{ k :
\Mcal_* \subseteq \Mcal^{(k)} \}$ is well defined and $k_{\min} \leq
m_{\max}^*$, for $m_{\max}^*$ defined in \eqref{eq:max_num_steps}. We
show that
\begin{equation}
\PP[\min_{k \in \{0, \ldots, k_{\min}-1\}} ({\rm BIC}(\Mcal^{(k)}) -
  {\rm BIC}(\Mcal^{(k+1)})) > 0] \rightarrow 1,
\end{equation}
so that $\PP[\hat s < k_{\min}] \rightarrow 0$ as $n \rightarrow
\infty$. We proceed by lower bounding the difference in the BIC scores
as
\begin{equation}
  \begin{aligned}
    {\rm BIC}(\Mcal^{(k)}) &- {\rm BIC}(\Mcal^{(k+1)}) = 
      \log \left( \frac{{\rm RSS}(\Mcal^{(k)})}{{\rm
        RSS}(\Mcal^{(k+1)})} \right)
      - \frac{\log(n) + 2 \log(p)}{n} \\
     & \geq \log\left( 1 + \frac{{\rm RSS}(\Mcal^{(k)}) - {\rm
          RSS}(\Mcal^{(k+1)})}{{\rm RSS}(\Mcal^{(k+1)})} \right)
      - 3n^{-1}\log(p),
  \end{aligned}
\end{equation}
where we have assumed $p > n$. Define the event $\Acal := \{
n^{-1}T^{-1} \sum_{t \in [T]}\norm{\yb_t}_2^2 \leq 2\}$. Note that
${\rm RSS}(\Mcal^{(k+1)}) \leq \sum_{t \in [T]}\norm{\yb_t}_2^2$, so
on the event $\Acal$ the difference in the BIC scores is lower bounded
as
\begin{equation}
\log(1 + 2n^{-1}T^{-1} \Delta(k)) - 3n^{-1}\log(p),
\end{equation}
where $\Delta(k)$ is defined in \eqref{eq:def_delta}. Using the fact
that $\log(1 + x) \geq \min(\log(2), 2^{-1}x)$ and the lower bound
from \eqref{eq:lower_bound_on_delta_final}, we have 
\begin{equation} \label{eq:lower_bound_bic}
{\rm BIC}(\Mcal^{(k)}) - {\rm BIC}(\Mcal^{(k+1)}) \geq \min( \log 2,
Cn^{-\delta_s - 2\delta_{\min}} ) - 3n^{-1}\log p,
\end{equation}
for some positive constant $C$. It is easy to check that $\log 2 -
3n^{-1}\log p > 0$ and $Cn^{-\delta_s - 2\delta_{\min}} - 3n^{-1}\log
p > 0$ under the conditions of the theorem. The lower bound in
\eqref{eq:lower_bound_bic} is uniform for $k \in \{0, \ldots,
k_{\min}\}$, so the proof is complete if we show that $\PP[\Acal]
\rightarrow 1$. But this easily follows from the tail bounds on the
central chi-squared random variable given in
Lemma~\ref{eq:chi_squared_tails}.

\subsection{Collection of known results}

In what follows, $C_1, C_2, \ldots$ will denote arbitrary positive constants.

The following result on the minimum eigenvalue of sub-matrices of the
covariance matrix $\hat \Sigmab$ is quite standard
(e.g. \cite{zhou09adaptive}, \cite{wang09forward} or
\cite{bickel09simultaneous}).
\begin{lemma} \label{lem:sparse_eigenvalue}
Let $\xb \sim \Ncal(0, \Sigmab)$ and $\hat \Sigmab = n^{-1} \sum_{i=1}^n
\xb_i\xb'_i$ be the empirical estimate from $n$ independent
realizations of $\xb$. Denote $\Sigmab = [\sigma_{ab}]$ and $\hat
\Sigmab = [\hat \sigma_{ab}]$. Assume $\phi_{min} \leq
\Lambda_{\min}(\Sigmab) \leq \Lambda_{\max}(\Sigmab) \leq
\phi_{\max}$. Then
\begin{equation} \label{eq:max_eigen_bound}
  \PP[\max_{\Mcal \subseteq [p], |\Mcal| < s} \Lambda_{\max}(\hat
  \Sigmab_{\Mcal}) \geq 2\phi_{\max}] \leq   
    C_1
    \exp(-C_2\frac{n}{s^2} + s\log p)
\end{equation}
and
\begin{equation} \label{eq:min_eigen_bound}
  \PP[\min_{\Mcal \subseteq [p], |\Mcal| < s} \Lambda_{\min}(\hat
  \Sigmab_{\Mcal}) \leq \phi_{\min}/2] \leq 
    C_3
    \exp(-C_4\frac{n}{s^2} + s\log p).
\end{equation}
\end{lemma}

The following tail bounds for the chi-squared distribution are taken
from~\cite{Laurent00adaptive}.
\begin{lemma} \label{eq:chi_squared_tails}
  Let $\chi^2_n$ be a central chi-squared r.v. with $n$
  degrees of freedom. For any positive $\epsilon$,
  \begin{align}
    \PP[\chi^2_n \geq n + 2\sqrt{n\epsilon} + 2\epsilon] & \leq
    \exp(-\epsilon) \\
    \PP[\chi^2_n \leq \epsilon - 2\sqrt{n\epsilon}] & \leq 
    \exp(-\epsilon).
  \end{align}
\end{lemma}

We also make use of the result taken from \cite{obozinski09high},
which bounds the maximum of a collection of chi-squared random variables.
\begin{lemma} \label{lem:max_chi_squared}
  Let $X_1, \ldots, X_m$ be {\it i.i.d.} central chi-squared r.v. with
  $n$ degrees of freedom. Then for any $\epsilon > n$,
  \begin{equation}
    \PP[\max_{i \in [m]} X_i \geq 2\epsilon] \leq m \exp(
    -\epsilon(1-2\sqrt{\frac{n}{\epsilon}})).
  \end{equation}
\end{lemma}

\subsection{Tables with simulation results}

\begin{adjustwidth}{-1in}{-1in}
  {\small
\begin{tabular}{c c c c c c c c c}
    
    \multicolumn{9}{c}{Simulation 1: $(n, p, s, T) = (500, 20000, 18, 500)$, $T_{\rm non-zero} = 500$}\\
    \hline\hline
    & Method name 
    & $\Mcal_* \subseteq \hat S$ 
    & Correct zeros 
    & Incorrect zeros
    & $\Mcal_* = \hat S$ 
    & $|\hat S|$ 
    & $\norm{\Bb - \hat \Bb}_2^2$ 
    & $R^2$ \\
    \hline
    \multicolumn{9}{c}{SNR = 15}\\
    \\
\multirow{5}{*}{\begin{sideways}Union\end{sideways} \begin{sideways} Support\end{sideways}}
& SIS-ALASSO & 100.0 & 100.0 & 0.0 & 10.0 & 20.2 & - & -\\
& ISIS-ALASSO & 100.0 & 100.0 & 0.0 & 18.0 & 19.6 & - & -\\
& OMP & 100.0 & 100.0 & 0.0 & 0.0 & 23.9 & - & -\\
& S-OMP & 100.0 & 100.0 & 0.0 & 100.0 & 18.0 & - & -\\
\hline

  \end{tabular}}
\end{adjustwidth}

\begin{adjustwidth}{-1in}{-1in}
  {\small
  \begin{tabular}{c c c c c c c c c}
    
    \multicolumn{9}{c}{Simulation 1: $(n, p, s, T) = (500, 20000, 18, 500)$, $T_{\rm non-zero} = 300$}\\
    \hline\hline
    & Method name 
    & $\Mcal_* \subseteq \hat S$ 
    & Correct zeros 
    & Incorrect zeros
    & $\Mcal_* = \hat S$ 
    & $|\hat S|$ 
    & $\norm{\Bb - \hat \Bb}_2^2$ 
    & $R^2$ \\
    \hline
    \multicolumn{9}{c}{SNR = 15}\\
    \\
\multirow{5}{*}{\begin{sideways}Union\end{sideways} \begin{sideways} Support\end{sideways}}
& SIS-ALASSO & 100.0 & 100.0 & 0.0 & 97.0 & 18.0 & - & -\\
& ISIS-ALASSO & 100.0 & 100.0 & 0.0 & 98.0 & 18.0 & - & -\\
& OMP & 100.0 & 100.0 & 0.0 & 0.0 & 23.0 & - & -\\
& S-OMP & 100.0 & 100.0 & 0.0 & 100.0 & 18.0 & - & -\\
\hline

  \end{tabular}}
\end{adjustwidth}

\begin{adjustwidth}{-1in}{-1in}
  {\small
  \begin{tabular}{c c c c c c c c c}
    
    \multicolumn{9}{c}{Simulation 1: $(n, p, s, T) = (500, 20000, 18, 500)$, $T_{\rm non-zero} = 100$}\\
    \hline\hline
    & Method name 
    & $\Mcal_* \subseteq \hat S$ 
    & Correct zeros 
    & Incorrect zeros
    & $\Mcal_* = \hat S$ 
    & $|\hat S|$ 
    & $\norm{\Bb - \hat \Bb}_2^2$ 
    & $R^2$ \\
    \hline

    \multicolumn{9}{c}{SNR = 15}\\
    \\
\multirow{5}{*}{\begin{sideways}Union\end{sideways} \begin{sideways} Support\end{sideways}}
& SIS-ALASSO & 100.0 & 100.0 & 0.0 & 100.0 & 18.0 & - & -\\
& ISIS-ALASSO & 100.0 & 100.0 & 0.0 & 100.0 & 18.0 & - & -\\
& OMP & 100.0 & 99.9 & 0.0 & 0.0 & 28.8 & - & -\\
& S-OMP & 100.0 & 100.0 & 0.0 & 100.0 & 18.0 & - & -\\
\hline

  \end{tabular}}
\end{adjustwidth}


\begin{adjustwidth}{-1in}{-1in}
  {\small
  \begin{tabular}{c c c c c c c c c}
    \multicolumn{9}{c}{Simulation 2.a: $(n, p, s, T) = (200, 5000, 10, 500)$, $T_{\rm non-zero} = 400$}\\
    \hline\hline
    & Method name 
    & $\Mcal_* \subseteq \hat S$ 
    & Correct zeros 
    & Incorrect zeros
    & $\Mcal_* = \hat S$ 
    & $|\hat S|$ 
    & $\norm{\Bb - \hat \Bb}_2^2$ 
    & $R^2$ \\
    \hline

    \multicolumn{9}{c}{SNR = 10}\\
    \\
\multirow{5}{*}{\begin{sideways}Union\end{sideways} \begin{sideways} Support\end{sideways}}
& SIS-ALASSO & 100.0 & 100.0 & 0.0 & 39.0 & 10.9 & - & -\\
& ISIS-ALASSO & 100.0 & 100.0 & 0.0 & 12.0 & 12.2 & - & -\\
& OMP & 100.0 & 99.8 & 0.0 & 0.0 & 21.6 & - & -\\
& S-OMP & 100.0 & 100.0 & 0.0 & 100.0 & 10.0 & - & -\\
\hline

  \end{tabular}}
\end{adjustwidth}

\begin{adjustwidth}{-1in}{-1in}
  {\small
  \begin{tabular}{c c c c c c c c c}    
    \multicolumn{9}{c}{Simulation 2.a: $(n, p, s, T) = (200, 5000, 10, 500)$, $T_{\rm non-zero} = 250$}\\
    \hline\hline
    & Method name 
    & $\Mcal_* \subseteq \hat S$ 
    & Correct zeros 
    & Incorrect zeros
    & $\Mcal_* = \hat S$ 
    & $|\hat S|$ 
    & $\norm{\Bb - \hat \Bb}_2^2$ 
    & $R^2$ \\
    \hline

    \multicolumn{9}{c}{SNR = 10}\\
    \\
\multirow{5}{*}{\begin{sideways}Union\end{sideways} \begin{sideways} Support\end{sideways}}
& SIS-ALASSO & 100.0 & 100.0 & 0.0 & 99.0 & 10.0 & - & -\\
& ISIS-ALASSO & 100.0 & 100.0 & 0.0 & 98.0 & 10.0 & - & -\\
& OMP & 100.0 & 99.8 & 0.0 & 0.0 & 19.9 & - & -\\
& S-OMP & 100.0 & 100.0 & 0.0 & 100.0 & 10.0 & - & -\\
\hline

  \end{tabular}}
\end{adjustwidth}

\begin{adjustwidth}{-1in}{-1in}
  {\small
  \begin{tabular}{c c c c c c c c c}
    
    \multicolumn{9}{c}{Simulation 2.a: $(n, p, s, T) = (200, 5000, 10, 500)$, $T_{\rm non-zero} = 100$}\\
    \hline\hline
    & Method name 
    & $\Mcal_* \subseteq \hat S$ 
    & Correct zeros 
    & Incorrect zeros
    & $\Mcal_* = \hat S$ 
    & $|\hat S|$ 
    & $\norm{\Bb - \hat \Bb}_2^2$ 
    & $R^2$ \\
    \hline

    \multicolumn{9}{c}{SNR = 10}\\
    \\
\multirow{5}{*}{\begin{sideways}Union\end{sideways} \begin{sideways} Support\end{sideways}}
& SIS-ALASSO & 100.0 & 100.0 & 0.0 & 100.0 & 10.0 & - & -\\
& ISIS-ALASSO & 100.0 & 100.0 & 0.0 & 100.0 & 10.0 & - & -\\
& OMP & 100.0 & 98.8 & 0.0 & 0.0 & 69.8 & - & -\\
& S-OMP & 100.0 & 100.0 & 0.0 & 100.0 & 10.0 & - & -\\
\hline

  \end{tabular}}
\end{adjustwidth}


\begin{adjustwidth}{-1in}{-1in}
  {\small
  \begin{tabular}{c c c c c c c c c}
    
    \multicolumn{9}{c}{Simulation 2.b: $(n, p, s, T) = (200, 5000, 10, 750)$, $T_{\rm non-zero} = 600$}\\
    \hline\hline
    & Method name 
    & $\Mcal_* \subseteq \hat S$ 
    & Correct zeros 
    & Incorrect zeros
    & $\Mcal_* = \hat S$ 
    & $|\hat S|$ 
    & $\norm{\Bb - \hat \Bb}_2^2$ 
    & $R^2$ \\
    \hline

    \multicolumn{9}{c}{SNR = 10}\\
    \\
\multirow{5}{*}{\begin{sideways}Union\end{sideways} \begin{sideways} Support\end{sideways}}
& SIS-ALASSO & 100.0 & 100.0 & 0.0 & 25.0 & 11.3 & - & -\\
& ISIS-ALASSO & 100.0 & 99.9 & 0.0 & 5.0 & 13.3 & - & -\\
& OMP & 100.0 & 99.7 & 0.0 & 0.0 & 26.6 & - & -\\
& S-OMP & 100.0 & 100.0 & 0.0 & 100.0 & 10.0 & - & -\\
\hline

  \end{tabular}}
\end{adjustwidth}

\begin{adjustwidth}{-1in}{-1in}
  {\small
  \begin{tabular}{c c c c c c c c c}
    
    \multicolumn{9}{c}{Simulation 2.b: $(n, p, s, T) = (200, 5000, 10, 750)$, $T_{\rm non-zero} = 375$}\\
    \hline\hline
    & Method name 
    & $\Mcal_* \subseteq \hat S$ 
    & Correct zeros 
    & Incorrect zeros
    & $\Mcal_* = \hat S$ 
    & $|\hat S|$ 
    & $\norm{\Bb - \hat \Bb}_2^2$ 
    & $R^2$ \\
    \hline

    \multicolumn{9}{c}{SNR = 10}\\
    \\
\multirow{5}{*}{\begin{sideways}Union\end{sideways} \begin{sideways} Support\end{sideways}}
& SIS-ALASSO & 100.0 & 100.0 & 0.0 & 99.0 & 10.0 & - & -\\
& ISIS-ALASSO & 100.0 & 100.0 & 0.0 & 93.0 & 10.1 & - & -\\
& OMP & 100.0 & 99.7 & 0.0 & 0.0 & 24.7 & - & -\\
& S-OMP & 100.0 & 100.0 & 0.0 & 100.0 & 10.0 & - & -\\
\hline

  \end{tabular}}
\end{adjustwidth}

\begin{adjustwidth}{-1in}{-1in}
  {\small
  \begin{tabular}{c c c c c c c c c}
    
    \multicolumn{9}{c}{Simulation 2.b: $(n, p, s, T) = (200, 5000, 10, 750)$, $T_{\rm non-zero} = 150$}\\
    \hline\hline
    & Method name 
    & $\Mcal_* \subseteq \hat S$ 
    & Correct zeros 
    & Incorrect zeros
    & $\Mcal_* = \hat S$ 
    & $|\hat S|$ 
    & $\norm{\Bb - \hat \Bb}_2^2$ 
    & $R^2$ \\
    \hline

    \multicolumn{9}{c}{SNR = 10}\\
    \\
\multirow{5}{*}{\begin{sideways}Union\end{sideways} \begin{sideways} Support\end{sideways}}
& SIS-ALASSO & 100.0 & 100.0 & 0.0 & 100.0 & 10.0 & - & -\\
& ISIS-ALASSO & 100.0 & 100.0 & 0.0 & 100.0 & 10.0 & - & -\\
& OMP & 100.0 & 98.0 & 0.0 & 0.0 & 108.5 & - & -\\
& S-OMP & 100.0 & 100.0 & 0.0 & 100.0 & 10.0 & - & -\\
\hline

  \end{tabular}}
\end{adjustwidth}


\begin{adjustwidth}{-1in}{-1in}
  {\small
  \begin{tabular}{c c c c c c c c c}
    
    \multicolumn{9}{c}{Simulation 2.c: $(n, p, s, T) = (200, 5000, 10, 1000)$, $T_{\rm non-zero} = 800$}\\
    \hline\hline
    & Method name 
    & $\Mcal_* \subseteq \hat S$ 
    & Correct zeros 
    & Incorrect zeros
    & $\Mcal_* = \hat S$ 
    & $|\hat S|$ 
    & $\norm{\Bb - \hat \Bb}_2^2$ 
    & $R^2$ \\
    \hline

    \multicolumn{9}{c}{SNR = 10}\\
    \\
\multirow{5}{*}{\begin{sideways}Union\end{sideways} \begin{sideways} Support\end{sideways}}
& SIS-ALASSO & 100.0 & 100.0 & 0.0 & 21.0 & 11.7 & - & -\\
& ISIS-ALASSO & 100.0 & 99.9 & 0.0 & 5.0 & 14.4 & - & -\\
& OMP & 100.0 & 99.6 & 0.0 & 0.0 & 32.0 & - & -\\
& S-OMP & 100.0 & 100.0 & 0.0 & 100.0 & 10.0 & - & -\\
\hline

  \end{tabular}}
\end{adjustwidth}

\begin{adjustwidth}{-1in}{-1in}
  {\small
  \begin{tabular}{c c c c c c c c c}
    
    \multicolumn{9}{c}{Simulation 2.c: $(n, p, s, T) = (200, 5000, 10, 1000)$, $T_{\rm non-zero} = 500$}\\
    \hline\hline
    & Method name 
    & $\Mcal_* \subseteq \hat S$ 
    & Correct zeros 
    & Incorrect zeros
    & $\Mcal_* = \hat S$ 
    & $|\hat S|$ 
    & $\norm{\Bb - \hat \Bb}_2^2$ 
    & $R^2$ \\
    \hline

    \multicolumn{9}{c}{SNR = 10}\\
    \\
\multirow{5}{*}{\begin{sideways}Union\end{sideways} \begin{sideways} Support\end{sideways}}
& SIS-ALASSO & 100.0 & 100.0 & 0.0 & 89.0 & 10.1 & - & -\\
& ISIS-ALASSO & 100.0 & 100.0 & 0.0 & 95.0 & 10.1 & - & -\\
& OMP & 100.0 & 99.6 & 0.0 & 0.0 & 29.1 & - & -\\
& S-OMP & 100.0 & 100.0 & 0.0 & 100.0 & 10.0 & - & -\\
\hline

  \end{tabular}}
\end{adjustwidth}

\begin{adjustwidth}{-1in}{-1in}
  {\small
  \begin{tabular}{c c c c c c c c c}
    
    \multicolumn{9}{c}{Simulation 2.c: $(n, p, s, T) = (200, 5000, 10, 1000)$, $T_{\rm non-zero} = 200$}\\
    \hline\hline
    & Method name 
    & $\Mcal_* \subseteq \hat S$ 
    & Correct zeros 
    & Incorrect zeros
    & $\Mcal_* = \hat S$ 
    & $|\hat S|$ 
    & $\norm{\Bb - \hat \Bb}_2^2$ 
    & $R^2$ \\
    \hline

    \multicolumn{9}{c}{SNR = 10}\\
    \\
\multirow{5}{*}{\begin{sideways}Union\end{sideways} \begin{sideways} Support\end{sideways}}
& SIS-ALASSO & 100.0 & 100.0 & 0.0 & 100.0 & 10.0 & - & -\\
& ISIS-ALASSO & 100.0 & 100.0 & 0.0 & 100.0 & 10.0 & - & -\\
& OMP & 100.0 & 97.4 & 0.0 & 0.0 & 139.6 & - & -\\
& S-OMP & 100.0 & 100.0 & 0.0 & 100.0 & 10.0 & - & -\\
\hline

  \end{tabular}}
\end{adjustwidth}


\begin{adjustwidth}{-1in}{-1in}
  {\small
  \begin{tabular}{c c c c c c c c c}
    
    \multicolumn{9}{c}{Simulation 3: $(n, p, s, T) = (100, 5000, 3,
      150)$, $T_{\rm non-zero} = 80$, $\rho = 0.2$ }\\
    \hline\hline
    & Method name 
    & $\Mcal_* \subseteq \hat S$ 
    & Correct zeros 
    & Incorrect zeros
    & $\Mcal_* = \hat S$ 
    & $|\hat S|$ 
    & $\norm{\Bb - \hat \Bb}_2^2$ 
    & $R^2$ \\
    \hline

    \multicolumn{9}{c}{SNR = 10}\\
    \\
\multirow{5}{*}{\begin{sideways}Union\end{sideways} \begin{sideways} Support\end{sideways}}
& SIS-ALASSO & 100.0 & 100.0 & 0.0 & 100.0 & 3.0 & - & -\\
& ISIS-ALASSO & 100.0 & 100.0 & 0.0 & 100.0 & 3.0 & - & -\\
& OMP & 100.0 & 99.8 & 0.0 & 0.0 & 20.0 & - & -\\
& S-OMP & 100.0 & 100.0 & 0.0 & 100.0 & 3.0 & - & -\\
\hline

  \end{tabular}}
\end{adjustwidth}

\begin{adjustwidth}{-1in}{-1in}
  {\small
  \begin{tabular}{c c c c c c c c c}
    
    \multicolumn{9}{c}{Simulation 3: $(n, p, s, T) = (100, 5000, 3,
      150)$, $T_{\rm non-zero} = 80$, $\rho = 0.5$ }\\
    \hline\hline
    & Method name 
    & $\Mcal_* \subseteq \hat S$ 
    & Correct zeros 
    & Incorrect zeros
    & $\Mcal_* = \hat S$ 
    & $|\hat S|$ 
    & $\norm{\Bb - \hat \Bb}_2^2$ 
    & $R^2$ \\
    \hline

    \multicolumn{9}{c}{SNR = 10}\\
    \\
\multirow{5}{*}{\begin{sideways}Union\end{sideways} \begin{sideways} Support\end{sideways}}
& SIS-ALASSO & 100.0 & 100.0 & 0.0 & 98.0 & 3.0 & - & -\\
& ISIS-ALASSO & 100.0 & 100.0 & 0.0 & 100.0 & 3.0 & - & -\\
& OMP & 100.0 & 99.8 & 0.0 & 0.0 & 20.1 & - & -\\
& S-OMP & 100.0 & 100.0 & 0.0 & 100.0 & 3.0 & - & -\\
\hline

  \end{tabular}}
\end{adjustwidth}

\begin{adjustwidth}{-1in}{-1in}
  {\small
  \begin{tabular}{c c c c c c c c c}
    
    \multicolumn{9}{c}{Simulation 3: $(n, p, s, T) = (100, 5000, 3,
      150)$, $T_{\rm non-zero} = 80$, $\rho = 0.7$ }\\
    \hline\hline
    & Method name 
    & $\Mcal_* \subseteq \hat S$ 
    & Correct zeros 
    & Incorrect zeros
    & $\Mcal_* = \hat S$ 
    & $|\hat S|$ 
    & $\norm{\Bb - \hat \Bb}_2^2$ 
    & $R^2$ \\
    \hline

    \multicolumn{9}{c}{SNR = 10}\\
    \\
\multirow{5}{*}{\begin{sideways}Union\end{sideways} \begin{sideways} Support\end{sideways}}
& SIS-ALASSO & 80.0 & 100.0 & 6.7 & 80.0 & 2.8 & - & -\\
& ISIS-ALASSO & 85.0 & 100.0 & 5.0 & 85.0 & 2.9 & - & -\\
& OMP & 100.0 & 99.8 & 0.0 & 0.0 & 22.0 & - & -\\
& S-OMP & 0.0 & 100.0 & 51.0 & 0.0 & 1.5 & - & -\\
\hline

  \end{tabular}}
\end{adjustwidth}


\begin{adjustwidth}{-1in}{-1in}
  {\small
  \begin{tabular}{c c c c c c c c c}
    
    \multicolumn{9}{c}{Simulation 4: $(n, p, s, T) = (150, 4000, 8,
      150)$, $T_{\rm non-zero} = 80$, $\rho = 0.2$ }\\
    \hline\hline
    & Method name 
    & $\Mcal_* \subseteq \hat S$ 
    & Correct zeros 
    & Incorrect zeros
    & $\Mcal_* = \hat S$ 
    & $|\hat S|$ 
    & $\norm{\Bb - \hat \Bb}_2^2$ 
    & $R^2$ \\
    \hline

    \multicolumn{9}{c}{SNR = 10}\\
    \\
\multirow{5}{*}{\begin{sideways}Union\end{sideways} \begin{sideways} Support\end{sideways}}
& SIS-ALASSO & 100.0 & 100.0 & 0.0 & 100.0 & 8.0 & - & -\\
& ISIS-ALASSO & 100.0 & 100.0 & 0.0 & 97.0 & 8.0 & - & -\\
& OMP & 100.0 & 99.9 & 0.0 & 2.0 & 11.7 & - & -\\
& S-OMP & 100.0 & 100.0 & 0.0 & 100.0 & 8.0 & - & -\\
\hline

  \end{tabular}}
\end{adjustwidth}

\begin{adjustwidth}{-1in}{-1in}
  {\small
  \begin{tabular}{c c c c c c c c c}
    
    \multicolumn{9}{c}{Simulation 4: $(n, p, s, T) = (150, 4000, 8,
      150)$, $T_{\rm non-zero} = 80$, $\rho = 0.5$ }\\
    \hline\hline
    & Method name 
    & $\Mcal_* \subseteq \hat S$ 
    & Correct zeros 
    & Incorrect zeros
    & $\Mcal_* = \hat S$ 
    & $|\hat S|$ 
    & $\norm{\Bb - \hat \Bb}_2^2$ 
    & $R^2$ \\
    \hline

    \multicolumn{9}{c}{SNR = 10}\\
    \\
\multirow{5}{*}{\begin{sideways}Union\end{sideways} \begin{sideways} Support\end{sideways}}
& SIS-ALASSO & 100.0 & 100.0 & 0.0 & 80.0 & 8.2 & - & -\\
& ISIS-ALASSO & 100.0 & 100.0 & 0.0 & 89.0 & 8.1 & - & -\\
& OMP & 100.0 & 99.9 & 0.0 & 2.0 & 11.9 & - & -\\
& S-OMP & 100.0 & 100.0 & 0.0 & 100.0 & 8.0 & - & -\\
\hline

  \end{tabular}}
\end{adjustwidth}


\begin{adjustwidth}{-1in}{-1in}
  {\small
  \begin{tabular}{c c c c c c c c c}
    
    \multicolumn{9}{c}{Simulation 5: $(n, p, s, T) = (200, 10000, 5,
      500)$, $T_{\rm non-zero} = 400$ }\\
    \hline\hline
    & Method name 
    & $\Mcal_* \subseteq \hat S$ 
    & Correct zeros 
    & Incorrect zeros
    & $\Mcal_* = \hat S$ 
    & $|\hat S|$ 
    & $\norm{\Bb - \hat \Bb}_2^2$ 
    & $R^2$ \\
    \hline

    \multicolumn{9}{c}{$\sigma$ = 1.5}\\
    \\
\multirow{5}{*}{\begin{sideways}Union\end{sideways} \begin{sideways} Support\end{sideways}}
& SIS-ALASSO & 53.0 & 99.6 & 9.4 & 0.0 & 41.1 & - & -\\
& ISIS-ALASSO & 100.0 & 99.8 & 0.0 & 0.0 & 28.1 & - & -\\
& OMP & 100.0 & 99.9 & 0.0 & 12.0 & 10.0 & - & -\\
& S-OMP & 100.0 & 100.0 & 0.0 & 44.0 & 5.6 & - & -\\
\hline

  \end{tabular}}
\end{adjustwidth}

\begin{adjustwidth}{-1in}{-1in}
  {\small
  \begin{tabular}{c c c c c c c c c}
    
    \multicolumn{9}{c}{Simulation 5: $(n, p, s, T) = (200, 10000, 5,
      500)$, $T_{\rm non-zero} = 250$ }\\
    \hline\hline
    & Method name 
    & $\Mcal_* \subseteq \hat S$ 
    & Correct zeros 
    & Incorrect zeros
    & $\Mcal_* = \hat S$ 
    & $|\hat S|$ 
    & $\norm{\Bb - \hat \Bb}_2^2$ 
    & $R^2$ \\
    \hline

    \multicolumn{9}{c}{$\sigma$ = 1.5}\\
    \\
\multirow{5}{*}{\begin{sideways}Union\end{sideways} \begin{sideways} Support\end{sideways}}
& SIS-ALASSO & 100.0 & 99.7 & 0.0 & 0.0 & 31.5 & - & -\\
& ISIS-ALASSO & 100.0 & 99.9 & 0.0 & 1.0 & 14.3 & - & -\\
& OMP & 100.0 & 99.7 & 0.0 & 0.0 & 30.8 & - & -\\
& S-OMP & 100.0 & 100.0 & 0.0 & 20.0 & 5.8 & - & -\\
\hline

  \end{tabular}}
\end{adjustwidth}

\begin{adjustwidth}{-1in}{-1in}
  {\small
  \begin{tabular}{c c c c c c c c c}
    
    \multicolumn{9}{c}{Simulation 5: $(n, p, s, T) = (200, 10000, 5,
      500)$, $T_{\rm non-zero} = 100$ }\\
    \hline\hline
    & Method name 
    & $\Mcal_* \subseteq \hat S$ 
    & Correct zeros 
    & Incorrect zeros
    & $\Mcal_* = \hat S$ 
    & $|\hat S|$ 
    & $\norm{\Bb - \hat \Bb}_2^2$ 
    & $R^2$ \\
    \hline

    \multicolumn{9}{c}{$\sigma$ = 1.5}\\
    \\
\multirow{5}{*}{\begin{sideways}Union\end{sideways} \begin{sideways} Support\end{sideways}}
& SIS-ALASSO & 100.0 & 99.9 & 0.0 & 1.0 & 10.9 & - & -\\
& ISIS-ALASSO & 100.0 & 100.0 & 0.0 & 56.0 & 5.7 & - & -\\
& OMP & 100.0 & 98.0 & 0.0 & 0.0 & 205.8 & - & -\\
& S-OMP & 99.0 & 100.0 & 0.2 & 4.0 & 6.0 & - & -\\
\hline

  \end{tabular}}
\end{adjustwidth}

\newpage

\bibliography{biblio}

\end{document}